\documentclass[10pt,twocolumn,letterpaper]{article}

\usepackage{iccv}
\usepackage[accsupp]{axessibility}
\usepackage{times}
\usepackage{epsfig}
\usepackage[utf8]{inputenc}
\usepackage{graphicx}
\usepackage{amsmath}
\usepackage{amssymb}
\usepackage{booktabs}
\usepackage{algorithm}
\usepackage{array}
\newcommand{\T}{\mathcal{T}}

\newcommand{\D}{\mathcal{D}}

\usepackage{dsfont}
\usepackage{algpseudocode}
\usepackage{algorithmicx}
\usepackage{makecell}
\usepackage{lipsum}
\usepackage{wrapfig}
\usepackage{bm}
\usepackage{multirow}
\usepackage{silence}
\usepackage{xcolor}

\usepackage{flushend}

\usepackage{pifont}

\usepackage{subcaption}
\usepackage[normalem]{ulem}
\useunder{\uline}{\ul}{}

\algnewcommand\algorithmicforeach{\textbf{for each}}
\algdef{S}[FOR]{ForEach}[1]{\algorithmicforeach\ #1\ \algorithmicdo}
\usepackage{enumitem}

\usepackage[pagebackref=true,breaklinks=true,letterpaper=true,colorlinks,bookmarks=false]{hyperref}

\iccvfinalcopy 

\usepackage[capitalize]{cleveref}
\crefname{section}{Sec.}{Secs.}
\Crefname{section}{Section}{Sections}
\Crefname{table}{Table}{Tables}
\crefname{table}{Tab.}{Tabs.}

\def\iccvPaperID{9120} 

\begin{document}

\title{Enhancing Modality-Agnostic Representations via Meta-learning for Brain Tumor Segmentation}


\author{\large
Aishik Konwer\textsuperscript{1},\hskip 1em
Xiaoling Hu\textsuperscript{1},\hskip 1em
Joseph Bae\textsuperscript{2}, \hskip 1em 
Xuan Xu\textsuperscript{1},\hskip 1em 
Chao Chen\textsuperscript{2},\hskip 1em
Prateek Prasanna\textsuperscript{2}
\\
\textsuperscript{1}Department of Computer Science, Stony Brook University \hskip 1em
\\
\textsuperscript{2}Department of Biomedical Informatics, Stony Brook University \hskip 1em
\\
{\tt\small \{akonwer,xiaolhu,xuaxu\}@cs.stonybrook.edu} \\
{\tt\small \{joseph.bae,chao.chen.1,prateek.prasanna\}@stonybrook.edu}  
}

\maketitle
\ificcvfinal\thispagestyle{empty}\fi

\begin{abstract}
   In medical vision, different imaging modalities provide complementary information. However, in practice, not all modalities may be available during inference or even training. Previous approaches, e.g., knowledge distillation or image synthesis, often assume the availability of full modalities for all subjects during training; this is unrealistic and impractical due to the variability in data collection across sites. We propose a novel approach to learn enhanced modality-agnostic representations by employing a meta-learning strategy in training, even when only limited full modality samples are available. Meta-learning enhances partial modality representations to full modality representations by meta-training on partial modality data and meta-testing on limited full modality samples. Additionally, we co-supervise this feature enrichment by introducing an auxiliary adversarial learning branch. More specifically, a missing modality detector is used as a discriminator to mimic the full modality setting. Our segmentation framework significantly outperforms state-of-the-art brain tumor segmentation techniques in missing modality scenarios.
\end{abstract}

\section{Introduction}
\label{sec:intro}

Multiple medical imaging modalities/protocols are required to provide complementary diagnostic cues to clinicians. 
For instance, 
multiple Magnetic Resonance Imaging (MRI) sequences (henceforth referred to as modalities), namely native T1, post-contrast T1 (T1c), T2-weighted (T2), and Fluid Attenuated Inversion Recovery (FLAIR) are used together to understand the underlying spatial complexity of brain tumors and their surroundings~\cite{mri1,mri2}. 
Deep learning approaches~\cite{seg1,seg2,seg3,seg4,seg5,seg6} have found great success in multimodal brain tumor segmentation and treatment response assessment. These conventional brain tumor segmentation methods perform well only when all four acquisition modalities are available as input (i.e., in the \textit{full modality setting}). However, in clinical practice, often only a subset of modalities are  available due to issues including image degradation, motion artifacts~\cite{artifact}, erroneous acquisition settings, and brief scan times. Hence it is crucial to develop robust modality-agnostic methods which can achieve state-of-the-art performance in \textit{missing modality settings}, i.e., when different modalities are unavailable during inference or even training.


\begin{figure}[t]
  \includegraphics[width=1.0\linewidth]{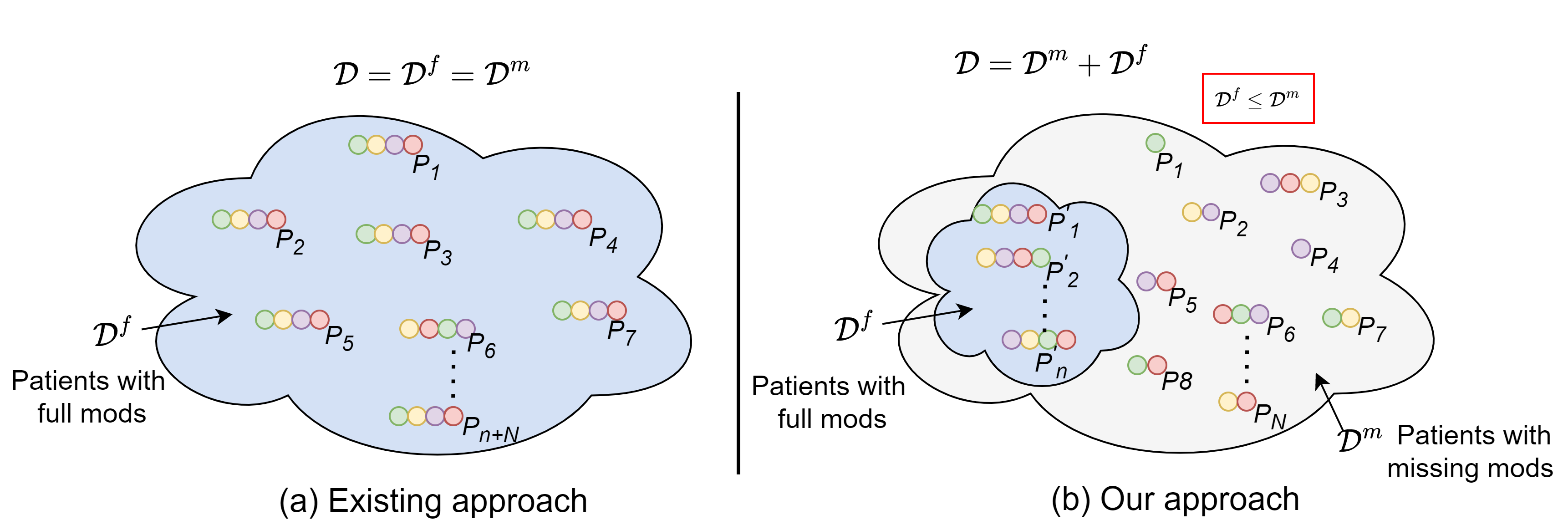}
    \caption{Comparison of the paradigms generally adopted by existing missing modality approaches (left) vs.~ours (right) for brain tumor segmentation. $N$ and $n$ refer to the number of subjects (patients) with partial and full modalities, respectively. Previous methods either utilize full modality data $\mathcal{D}^f$ for all subjects or simulate partial modality data $\mathcal{D}^m$ from $\mathcal{D}^f$. On the contrary, our approach works in a limited full modality setting, i.e.,  $|\mathcal{D}^f|\le|\mathcal{D}^m|$.}
            \vspace{-.1in}
    \label{missing_modality}

\end{figure}

Recently, a plethora of works 
has been proposed to address missing modality scenarios for brain tumor segmentation. Two major categories include: 1) Knowledge distillation: These methods ~\cite{kd1,kd2,kd3,kd4,kd5} learn privileged information from a teacher network trained on full modality data, i.e., data with all modalities available.
2) Image synthesis: Several works~\cite{gen1,gen2,gen3,gen4,gen5} train generative models to synthesize images of the missing modalities. The synthesized ``full modality'' images are used for segmentation. One major issue is that both categories of methods require full modality data for all subjects in the training set (see Fig.~\ref{missing_modality}a), either to train the teacher or the generator. This can be very unrealistic; in real-world applications, most studies only have very limited full modality data, far from sufficient for training. In this paper, we focus on a more realistic setting: most training data is only partial modality data, i.e., having a few modalities missing. We ask the following question: \emph{How do we efficiently learn from a large amount of partial modality data and a small amount of full modality data (see Fig.~\ref{missing_modality}b)?}


Another category recently rising in popularity is Shared Latent Space modeling~\cite{hemis,hved, urn,chartsias,rfnet,mmformer,mfi,d2net,gpvae,moddrop++, latent, robust}. These methods learn a shared latent representation from partial modality data. However, the quality of the learned representation can be limited by the heterogeneity of available modalities. The learned representation will be biased towards the most frequently available modalities and essentially overlook minority modalities (i.e., modalities that appear less frequently in training). This will inevitably lead to sub-optimal performance on test data, especially with minority modalities. To compensate for this undesirable bias, these methods often resort to segmenting all modalities individually from the shared representation, ultimately requiring full modality for all cases during training. 

These observations, further summarized in Tab.~\ref{tabadvcomp}, motivate us to design a modality-agnostic method that can fully utilize partial modality data. Through the usage of the meta-learning strategy, our method  
learns enriched shared representations that are generalizable and not biased towards more frequent modalities, even with limited full modality data.

\setlength{\tabcolsep}{2.5pt}
\begin{table}[hbtp]
\centering
\scriptsize
\begin{tabular}{c|cc}
\hline
\multicolumn{1}{c|}{Category}   & \multicolumn{1}{c}{Can handle limited FM?} & \multicolumn{1}{c}{Learns Unbiased mapping?} \\ \cline{2-3} \hline
KD~\cite{kd2, kd3, kd4, kd5}                                      & \multicolumn{1}{c}{N}         &  \multicolumn{1}{c}{Y}         \\
                                           GAN~\cite{gen1, gen2, gen3, gen4, gen5}     &  \multicolumn{1}{c}{N}         &  \multicolumn{1}{c}{Y}        \\
                                            Shared (others)~\cite{robust, rfnet, mmformer}    &  \multicolumn{1}{c}{N}         &  \multicolumn{1}{c}{N}           \\
                                            SMIL~\cite{smil}    &  \multicolumn{1}{c}{Y}         &  \multicolumn{1}{c}{N}            \\
                                Shared (Ours)       &  \multicolumn{1}{c}{Y}         &  \multicolumn{1}{c}{Y}             \\\hline
\end{tabular}
\caption{Advantages of our approach over existing frameworks. We are able to train in a limited full modality (FM) setting (with $\le$ 50\% FM samples), and learn an unbiased mapping that is unaffected by the proportion of any given modality in training.}
\label{tabadvcomp}
\end{table}


\begin{figure}[t]
  \includegraphics[width=1.0\linewidth]{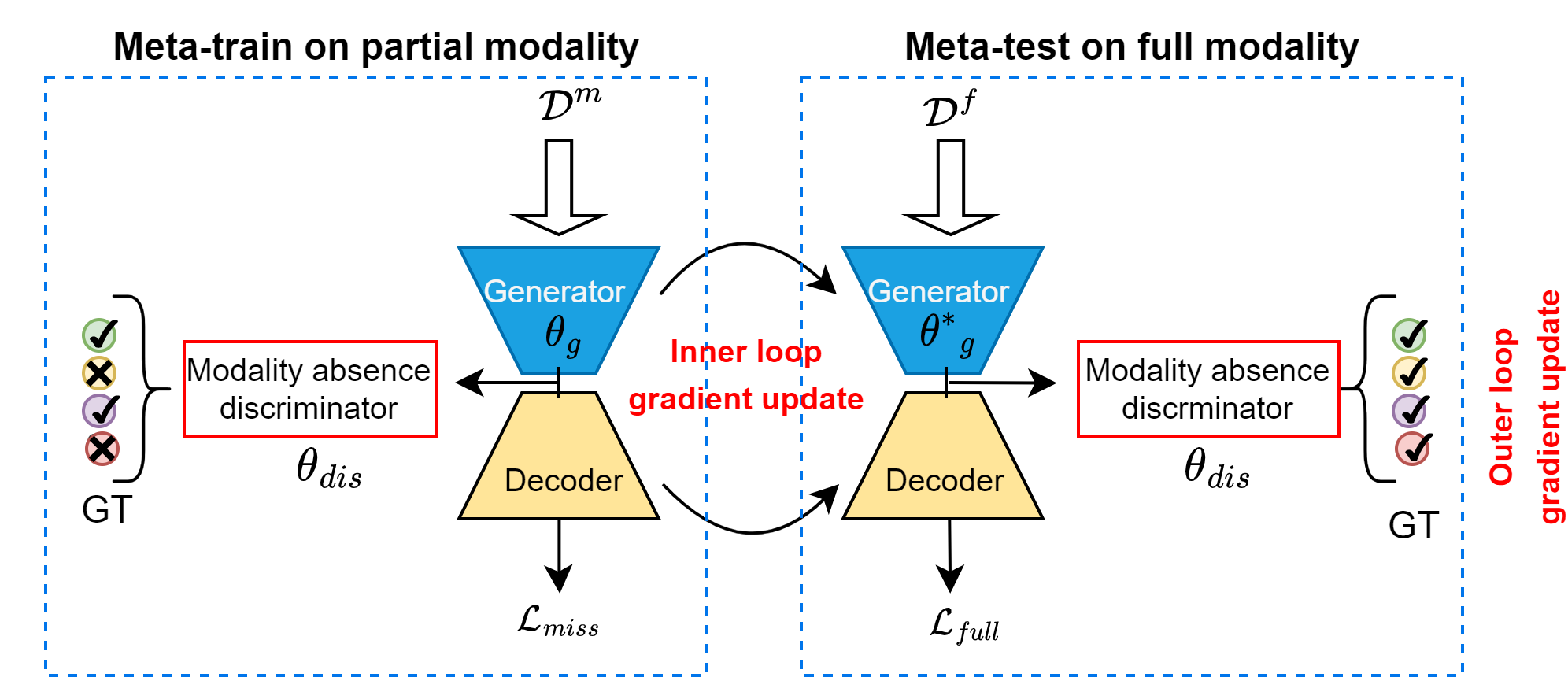}
    \caption{Framework overview. $\mathcal{D}^m$ (partial modality) and $\mathcal{D}^f$ (full modality) are used as inputs for encoder-decoder networks in the meta-train and meta-test phase, respectively. Partial modality representations are adapted to the full modality domain via:~1) meta-optimization of gradients in both data, and 2) adversarial learning based on predictions by a modality absence classifier.}
        \vspace{-.1in}
    \label{meta_learning}
\end{figure}


Our core idea is based on the meta-learning technique \cite{meta1}. Meta-learning provides an effective framework to learn to perform multiple tasks in a mutually beneficial manner. We consider segmentation with each partial modality input combination as a different task, yielding $2^M-1$ meta-tasks for $M$ modalities. By learning all meta-tasks in parallel, meta-learning ensures the network generates modality-agnostic representations. Thanks to meta-learning, tasks depending on rare modalities can be significantly improved even with limited training data. This maximally mitigates the bias against rare modalities. Meanwhile, we propose using a small amount of full modality data only during meta-testing. Meta-testing is introduced as an intermediate step in meta-learning to boost the generalization performance of the model across different tasks. Using full modality data, albeit limited, in meta-test can maximally leverage such data to enhance the representation quality of the model. This innovative meta-learning design ensures we learn with a large amount of partial modality data and only a small amount of full modality data, with negligible partial modality bias.

Recently a meta-learning approach~\cite{smil} performed classification with missing modalities. They predict the prior weights of modalities via a feature reconstruction network, the quality of which is indirectly dependent on the number of full modality samples. This method is unsuitable for our segmentation framework since conventional approaches (PCA~\cite{pca}, K-Means~\cite{kmeans}) cannot be used to cluster the priors in a high dimensional latent space. Moreover, ~\cite{smil} deals with only two input modalities and considers them individually as meta-tasks, while we construct a heterogeneous task distribution with different combinations of inputs respecting the heterogeneity settings of real-world data.

We also employ a novel adversarial learning technique that further enhances the quality of the generated shared latent space representation. Previous GAN-based approaches~\cite{gen1} reconstruct the missing modalities in image space; this leads to the impractical requirement of full modality as ground truth for training. Our task is achieved in latent space by designing the discriminator as a multi-label classifier. The discriminator predicts the presence/absence of modalities from the fused latent representation performing a binary classification for each modality. Our ultimate goal is to hallucinate the full modality representation from the hetero-modal feature space. Note that due to the hetero-modal nature of the data, the number of available modalities can vary dramatically across subjects. To address this, we utilized a channel-attention weighted fusion module that can accept a varying number of representations as input but generates a single fused output.

Overall, our contributions can be summarized as follows: 
\begin{itemize}
    \item We propose a meta-learning paradigm to train with hybrid data (partial and full modalities) and also enhance the learned partial modality representations to mimic a full modality representation. This is accomplished by meta-training on partial modality data while finetuning on limited full modality data during the meta-test. Such a training strategy overcomes the over-reliance on full modality data, as well as succeeds in learning an unbiased representation for all missing situations.
    \item We introduce a 
    novel adversarial learning strategy to further enrich the shared representations in the latent space. It differs from other generative approaches that synthesize missing images and demand full modality ground truths for training. Our approach does not necessitate reconstructing missing modality images. 

\end{itemize}
\section{Related Work}

\noindent\textbf{Segmentation with missing modalities.}
Incomplete data is a long-standing issue in computer vision; it particularly has significant implications in medical vision. 
Due to privacy concerns and budget constraints, one or more modalities (audio/visual/text)~\cite{privacy,emotion1,visual,clinic1,moieccv} of a given sample may not be available.
In this work, we focus on partial medical imaging modalities for brain tumor segmentation.
Existing methods on complete multi-modal brain tumor segmentation~\cite{seg1,seg2,seg3,seg4,seg5,seg6} perform poorly in realistic hetero-modal settings. 

Researchers have broadly used three techniques to perform brain tumor segmentation from missing modalities including 
\textit{knowledge distillation} (KD) ~\cite{kd1,kd2,kd3,kd4,kd5}, \textit{generative modeling}~\cite{gen1,gen2,gen3,gen4,gen5} and \textit{shared representation learning}~\cite{hemis,hved,urn,chartsias,rfnet,mmformer,mfi,d2net,gpvae,moddrop++,latent,robust}. 
ACN~\cite{kd3} trains a separate teacher-student pipeline for each subset of modalities. 
Among the generative models, MM-GAN~\cite{gen1}
uses a U-Net to impute missing modalities 
while a PatchGAN learns to discriminate between real and synthesized inputs. 
A major drawback of KD and GAN-based approaches is their inability to perform well when all modalities are not present for a subject during training. 
Moreover, the unstable and non-converging nature of a 3D generator may lead to degraded quality of synthesized images, eventually affecting downstream performance. Our method belongs to the third category, i.e., shared latent space models. In~\cite{hemis,hved}, the authors compute variational statistics to construct unified representations for segmentation. 
Multi-source information is modeled using a correlation constraint~\cite{latent} or region-aware fusion blocks~\cite{rfnet} to encode shared representations. Recent frameworks~\cite{mfi,mmformer} in this genre advocate for exploiting intra/inter-modality relations through graph and transformer-based modeling. Such approaches usually lack flexibility for adaptation to all missing scenarios. They yield sub-optimal performance due to failure in the retrieval of discriminative features generally existing in full modality data. Furthermore, these approaches can learn biased mappings among the available modalities leading to poor generalizability for modalities not encountered in training.

\begin{figure*}[hbtp]
  \centering
  \includegraphics[width=1.0\linewidth]{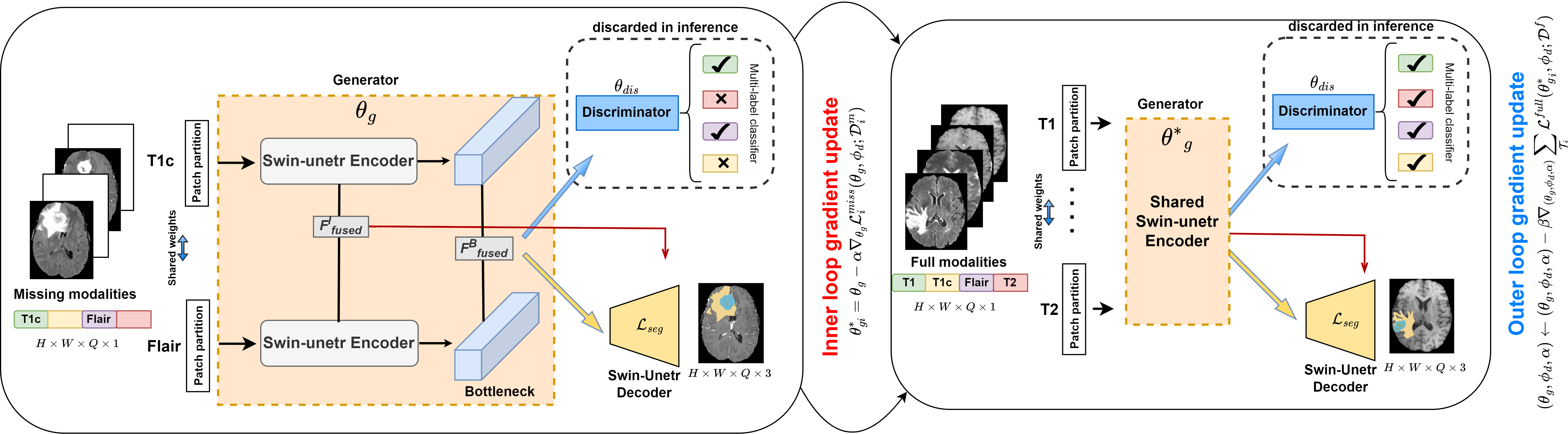}
    \caption{Illustration of the proposed framework. Available or full set of modalities are passed through a shared generator in the meta-train and meta-test stages respectively. The aggregation module helps to obtain a fused representation from five different levels (level $l$ and bottleneck are indicated here). Next, only the bottleneck embedding is used by the discriminator to predict which modalities are present at the input. All five fused embeddings are used by the segmentation decoder. Inner and outer loop gradient updates refer to the losses calculated in the meta-train and meta-test stages on partial modality and full modality data, respectively.}
    \label{framework}
\end{figure*}

\noindent\textbf{Meta-learning.}
Meta-learning algorithms~\cite{meta1,meta2,meta3} are inspired by human perception of new tasks. 
Optimization-based meta-learning techniques~\cite{meta1,meta2,meta4} have gained popularity since they can easily encode prior information through an optimization process. Model-agnostic meta-learning (MAML)~\cite{metric} is the most commonly used algorithm under this category, due to its flexible application to any network trained through gradient descent. 
Researchers have widely adopted MAML frameworks to generalize a model to new tasks, unseen domains, and enriching input features in multimodal scenarios~\cite{mamldomain,multimeta}. SMIL~\cite{smil} introduces a Bayesian MAML framework that attains comparable classification performance across both partial and full modality data. However, their approach requires prior reconstruction of the missing modalities. HetMAML~\cite{hetmaml} 
can handle heterogeneous task distributions, i.e. different modality combinations for input space but fails to attain generalizable performance across partial and full modalities. Inspired by the above two approaches, we propose a modality-agnostic architecture that can not only accept hetero-modal inputs but also enhance their representations with the additional information present in a full set of modalities. This leads to better segmentation performance for any hetero-modal input instance.

\noindent\textbf{Domain adaptation.}
Domain adaptation refers to the training of a neural network to jointly generate both discriminative and domain-invariant features in order to model different source and target data distributions~\cite{domain1,domain2,domain3,domain4,tcm}. 
Authors in~\cite{domain1} leverage an auxiliary domain classifier to address the domain shift. Inspired by this approach, we design our discriminator as a modality absence predictor. Similar to Sharma et al.~\cite{gen1}, we feed our discriminator with the correct modality code as ground truth, while the generator is provided an `all-one' full modality code impersonating the presence of all modalities. In an attempt to fool the discriminator, the generator learns to always mimic full modality representations, irrespective of the available inputs. This results in enhanced representations that boost downstream performance in missing modality situations.

\section{Methodology}
\noindent\textbf{Overview.} Given heterogeneous modalities as input, our goal is to build a modality-agnostic framework that can be robust to missing modality scenarios, and achieve performances comparable to a full modality setting. 
We have limited access to full modality data during training; this simulates a practical clinical scenario where brain tumor segmentation may need to be performed with partial modalities. To address this data-scarce situation, we aim to establish a mapping between partial and full modality representations. Our proposed approach is shown in Fig.~\ref{framework}. Meta-learning has been shown to be an efficient computational paradigm in dealing with heterogeneous training data~\cite{hetmaml}, or conducting feature adaptation between different domains~\cite{smil}. To this end, we adapt model-agnostic meta-learning to leverage information from both partial and full modality data. 
This strategy is elaborated in Sec.~\ref{sec:meta_learning}. We also want to further enrich encoded representations obtained from available modalities, with the supplemental information contained in full modality representations. More specifically, we propose a novel adversarial learning technique introducing a discriminator that acts as a modality absence classifier. A detailed description is provided in Sec.~\ref{sec:adversarial}. Because we need to generate a common fused representation for each hetero-modal input combination, our architecture incorporates a simple and elegant feature aggregation module (see Sec.~\ref{sec:aggregation}). 

\subsection{ Meta-Learning for Feature Adaptation }
\label{sec:meta_learning}
Suppose we have a total of $M$ MRI modalities as inputs for a patient. To simulate a real-world clinical scenario where full modalities may only be available for a fraction of subjects, some modalities are dropped for each patient during training. This training paradigm ensures the model becomes more robust to missing scenarios at inference. Thus we construct a heterogeneous task distribution $P(\T)$ that is a collection of $k$ task distributions: $P(\T^1),P(\T^2),...,P(\T^k)$. Each such distribution $P(\T^i)$ has a distinct input feature space related to a specific subset of modalities. We however exclude the full modality subset from the task distribution due to its utilization in meta-testing, as explained in the following paragraph. Overall, $k$ types of task instances can be sampled from $P(\T)$, where $k={2}^M-2$.

\setlength{\textfloatsep}{0pt}
\begin{algorithm}[H]
	\caption{Modality-Agnostic Meta-Learning}
	\label{algo:train}
	\begin{algorithmic}[1]
        \State \textbf{Input}: Training dataset $\D$ is divided into two cohorts of subjects with partial/missing and full modalities respectively $\mathcal{D}=\{\mathcal{D}^{m}, \mathcal{D}^{f}\}$; $\beta$ is the learning rate. 
        \State \textbf{Initialise}: Initialise $\theta_g, \phi_d =\{\theta_{dis}, \theta_{dec}\}, \alpha$
        \State \textbf{Output}: Optimized meta-parameters $\{\theta_g, \phi_d, \alpha\}$
        \While {not converged}
            \State Sample a batch of tasks $\mathcal{T}^i \sim \{P(\T)\} $ 
            \ForEach {task $\mathcal{T}^i$}
                \State Evaluate inner loop loss: $\mathcal{L}^{miss}_{i}(\theta_g, \phi_d; \mathcal{D}^{m}_i)$
                \State Adapt: $\theta^{*}_{g_i} = \theta_g - \alpha \nabla_{\theta_g} \mathcal{L}^{miss}_{i}(\theta_g, \phi_d; \mathcal{D}^{m}_i)$
                \State Compute outer loop loss: $\mathcal{L}^{full}({\theta^{*}_g}_i, \phi_d; \mathcal{D}^{f})$
             \EndFor
             \State Update meta-parameters:  $(\theta_g, \phi_d, \alpha) \leftarrow (\theta_g, \phi_d, \alpha)  -\indent \beta \nabla_{(\theta_g, \phi_d, \alpha)}  \sum_{\mathcal{T}_i} \mathcal{L}^{full} ({\theta^{*}_g}_i, \phi_d ; \mathcal{D}^{f})$
        \EndWhile
	\end{algorithmic} 
 \end{algorithm}

Formally, we have a hetero-modal training dataset $\D$ which we divide into two cohorts of subjects $\{\D^m,\D^f\}$ containing partial and full modalities, respectively. The goal is to effectively learn from both types of data.
We construct a batch of subjects $\D^{m}_i$ corresponding to each $P(\T^i)$. The pair of subjects and their corresponding task remains fixed over all epochs. Only the modalities which are not included in a task get dropped for that particular subject. 

Shared encoders are used along with a fusion module to produce a modality-agnostic representation. In our case, both encoder and fusion modules jointly constitute the generator $E_{\theta_g}$ (parameterized by $\theta_g$). An MLP-based classifier network, parameterized by $\theta_{dis}$, is employed as a discriminator as explained in Sec.~\ref{sec:adversarial}. For clarity, parameters of the discriminator and decoder network, $\{\theta_{dis}, \theta_{dec}\}$, are collectively symbolized as $\phi_d$. Our aim is to obtain an optimal generator parameter $\theta_g$ through task-wise training on $\D^{m}_{i}$ by reducing the inner loop objective $\mathcal{L}^{miss}_{i}$.
\begin{equation}
    \theta^{*}_{gi} = \theta_g - \alpha \nabla_{\theta_g} \mathcal{L}^{miss}_{i}(\theta_g, \phi_d; \mathcal{D}^{m}_i), 
\end{equation}
where $\alpha$ is a learnable rate for inner-level optimization. The optimized model is expected to perform better on $\D^f$.
The goal of the updated framework is to accomplish the outer loop objective $\mathcal{L}^{full}$ across all sampled tasks:
\begin{equation}
    \min_{\theta_g, \phi_d} \sum_{\mathcal{T}_i} \mathcal{L}^{full}({\theta^{*}_g}_i, \phi_d; \mathcal{D}^{f}). 
\end{equation}
Both the inner and outer loop losses are kept the same, referring to the generator and discriminator losses, $\mathcal{L}_{E}$ and $\mathcal{L}_{dis}$. By forcing the partial modality trained model to perform well on full modality data, we implicitly target the recovery of relevant information for better segmentation in missing modality scenarios. This partial to full modality mapping in feature space, is further strengthened by the introduction of a domain-adaptation inspired feature enrichment module (Details in Sec.~\ref{sec:adversarial}). All three meta parameters $(\theta_g, \phi_d, \alpha)$ are henceforth meta-updated by averaging gradients of outer loop loss over a meta-batch of tasks.
\begin{equation}
    (\theta_g, \phi_d, \alpha) \leftarrow  
    (\theta_g, \phi_d, \alpha) - \beta \nabla_{(\theta_g, \phi_d, \alpha)}  \sum_{\mathcal{T}_i} \mathcal{L}^{full} ({\theta^{*}_g}_i, \phi_d ; \mathcal{D}^{f}).
\end{equation}
Thus during meta-training, the model tunes its initialization parameter to achieve improved generalizability across all missing modality tasks. During meta-test, by finetuning with full modality data, we map the learned feature representations to the full modality space. Different from MAML, the pretrained model is directly evaluated on datasets where subjects contain a fixed subset of modalities (one of the tasks $\T^i$ already encountered in meta-training) at inference. The training process is summarized in Alg.~\ref{algo:train}.

\subsection{Adversarial Feature Enrichment}
\label{sec:adversarial}
Considering that full modality data contains richer information, we enforce encoder outputs to mimic full representations, irrespective of the limited input combination. The modality encoders and fusion module can be collectively considered as a shared generator $E$. We introduce an MLP-based multi-label classifier as our discriminator $D$.

The objective of $D$ is to predict the absence/presence of modalities from the fused embedding $\mathbf{F}^{B}_{fused}$ at the bottleneck level. $D$ utilizes Binary Cross-Entropy loss $\mathcal{L_{BCE}}$, and sigmoid activation to output $M$ binary predictions $\hat{d}$, denoting whether a modality is available or not. While calculating the discriminator loss $\mathcal{L}_{dis}$ indicated below, the ground truth variable $T_{real}$ is a vector of size $M$ which reflects the true combination of modalities available at input for that iteration. For example, assuming that $M=4$, and only first two modalities are available, $T_{real}=\{1,1,0,0\}$. 
\begin{equation}
    \mathcal{L}_{dis} = \sum_{z = 1}^{\mathcal{D}^m + \mathcal{D}^f}\mathcal{L_{BCE}}(\hat{d}_{z},T_{{real}_{z}}).
\end{equation}
The generator loss is a combination of segmentation loss and an adversarial loss used to train the generator to fool the discriminator. We consider a dummy ground truth variable $T_{dummy}$. In order to encourage the
generator to encode representations that confuse or “fool” the discriminator into inferring that all modalities are present, we set $T_{dummy}=\{1,1,1,1\}$, masquerading all generated representations as full modality representations. Thus $D$ pushes the
generator $E$ to agnostically produce full modality representations. 
\begin{equation}
    \mathcal{L}_E = \lambda_1\mathcal{L}_{seg} +  \lambda_{2}\sum_{z = 1}^{\mathcal{D}^m + \mathcal{D}^f}\mathcal{L_{BCE}}(\hat{d}_{z},T_{{dummy}_{z}}).
\end{equation}

\subsection{Modality-Agnostic Feature Aggregation}
\label{sec:aggregation}
We aim to utilize multiple modalities (which vary in number per patient) and derive a common fused representation. Individual encoders $E_1,E_2,...,E_n$ having shared parameters are trained to extract features from each of the $n$ available patient-specific modalities, where $1\le n \le M$. These features $\mathbf{F}^l_1,\mathbf{F}^l_2,...,\mathbf{F}^l_n$ obtained from the corresponding levels $(l)$ of each encoder are passed into a feature aggregation module. 

\begin{figure}[hbtp]
  \includegraphics[width=0.95\linewidth]{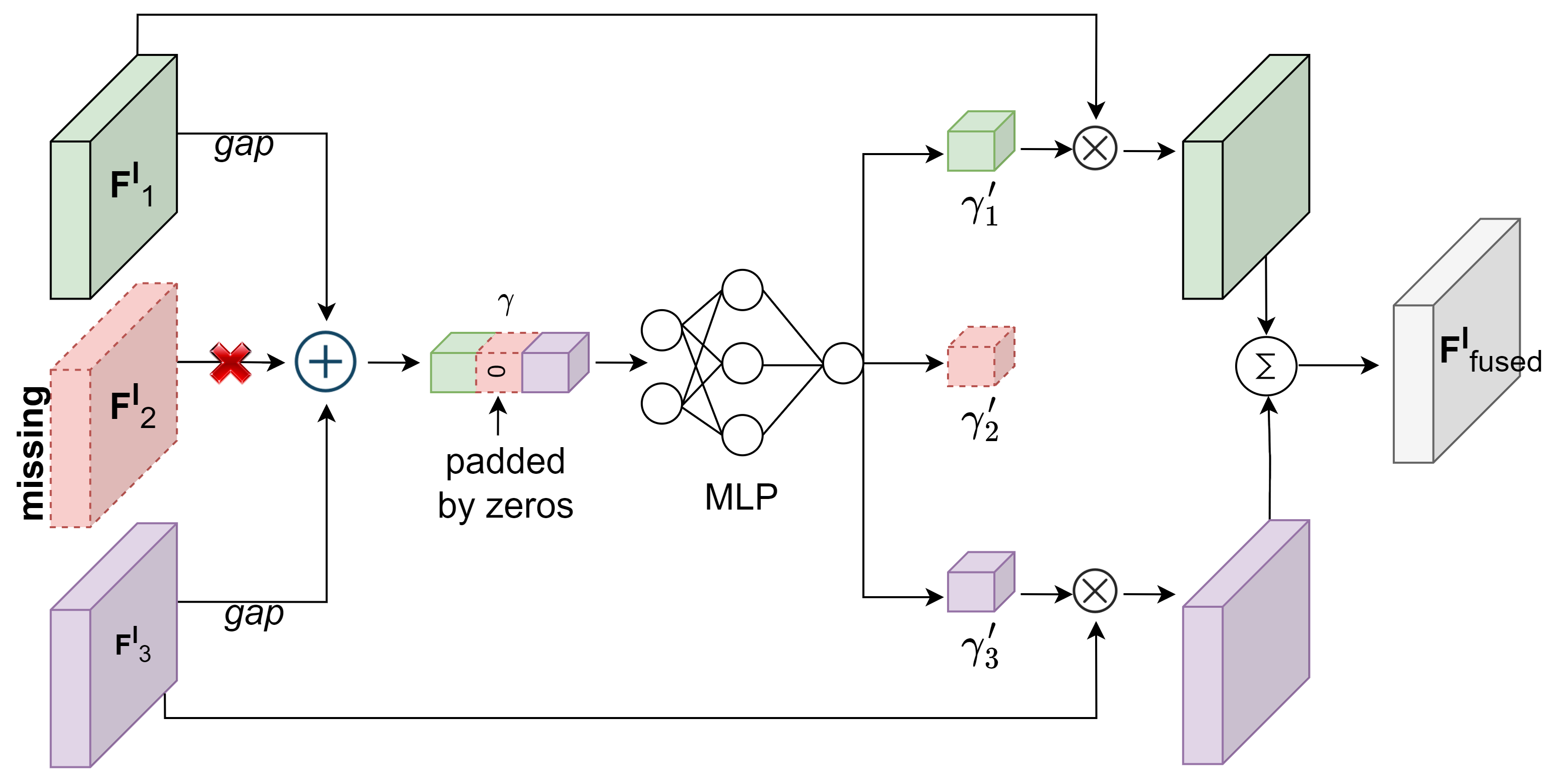}
    \caption{Illustration of the feature aggregation module. Modality $\mathbf{F}^l_2$ is missing. $\mathbf{F}^l_1$ and $\mathbf{F}^l_3$ are passed through global average pooling (GAP) operation and eventually fed into an MLP to generate the shared representation $\mathbf{F}^l_{fused}$. 
    }
    \label{fusion}
\end{figure}

The individual encoded representations undergo a Global Average Pooling operation and are subsequently concatenated to form a $M$-dimensional vector $\gamma$. This is achieved by imputing zeros in the channel information of ($M-n$) missing modalities. $\gamma$ is mapped to the channel weights
of $M$ modality features through a multi-layer perceptron $(MLP)$ and sigmoid activation function, $\sigma$. These modality-specific weights multiplied with the corresponding features give rise to the aggregated representation, $\mathbf{F}^{l}_{fused}$ (in Fig.~\ref{fusion}), which is eventually used as input to the decoder for segmentation. Our aggregation module exploits the correlation among available modality representations to create a unified feature that best describes the tumor characteristics of a subject. Detailed explanations with equations regarding this module can be found in the supplementary (Sec. 12).

We adopt a Swin-UNETR~\cite{seg6} architecture that employs soft Dice loss~\cite{dice} to perform voxel-wise semantic segmentation. The segmentation loss function $\mathcal{L}_{Seg}$ is defined as follows:
\begin{equation*}
\mathcal{L}_{seg}(G,P) = 1-\frac{2}{V}\sum_{v=1}^{V}\frac{\sum_{u=1}^{U}G_{u,v}P_{u,v} }{\sum_{u=1}^{U}G^{2}_{u,v}+ \sum_{u=1}^{U}P^{2}_{u,v}}.
\end{equation*}
where $V$ is the number of classes and $U$ is the number of voxels. $P_{u,v}$ and $G_{u,v}$ refer to the predicted output and one-hot encoded ground truth for class $v$ at voxel $u$, respectively.

\setlength{\tabcolsep}{2.5pt}
\begin{table*}[t]
\footnotesize
\begin{tabular}{c|c|c|c|c|c|c|c|c|c|c|c|c|c|c|c|c|c}
\hline
\multicolumn{1}{c|}{\multirow{4}{*}{M}} & FLAIR & $\circ$ & $\circ$ & $\circ$ & $\bullet$ & $\circ$  & $\circ$  & $\bullet$ & $\circ$ & $\bullet$ & $\bullet$ &$\bullet$  & $\bullet$ & $\bullet$ & $\circ$ & $\bullet$ & \multirow{4}{*}{\begin{tabular}[c]{@{}l@{}} Avg $\pm$ std,\\ p-value $(10^{-2})$ \end{tabular}}
\\ 
\multicolumn{1}{c|}{}                   & T1    &  $\circ$ & $\circ$  &$\bullet$  & $\circ$ & $\circ$ & $\bullet$ & $\bullet$ & $\bullet$ & $\circ$ & $\circ$ & $\bullet$ & $\bullet$ & $\circ$  & $\bullet$ & $\bullet$ &                           \\ 
\multicolumn{1}{c|}{}                   & T1c  & $\circ$ &$\bullet$ & $\circ$ & $\circ$ &$\bullet$  &$\bullet$  & $\circ$ & $\circ$ & $\circ$ & $\bullet$ & $\bullet$ & $\circ$ & $\bullet$ & $\bullet$ & $\bullet$ &                             \\
\multicolumn{1}{c|}{}                   & T2    & $\bullet$ & $\circ$ & $\circ$ & $\circ$ & $\bullet$ & $\circ$ & $\circ$ &$\bullet$  &$\bullet$  & $\circ$ & $\circ$ & $\bullet$ & $\bullet$ & $\bullet$ & $\bullet$ &                            \\ \hline
\multirow{7}{*}{{WT}}                 &                     HeMIS\cite{hemis}     & 79.85  & 60.32 & 55.76 & 66.20 & 81.63 & 65.39 & 75.41 & 81.70 & 82.56 & 76.25 & 79.82 & 84.58 & 86.27 & 82.74 & 85.06 & 76.23$\pm$9.66, $0.05^\ast$                             \\ 
                                         & U-HVED\cite{hved}     & 81.06 & 58.74 & 52.37 & 82.65 & 80.88 & 66.21 & 83.70 & 82.83 & 86.44 & 84.92 & 86.33 & 87.56 & 87.84 & 83.47 & 88.25 & 79.55$\pm$11.14, $2.10^\ast$ 
                                          \\ 
                                         & D2-Net\cite{d2net}     & 76.58  &  43.79 & 19.43 & 85.06 & 84.62 & 65.37 & 86.18 & 82.56 & 86.35 &  87.94 & 87.31 & 88.59 & 89.12 &  83.78 & 88.94 & 77.04$\pm$19.95, 6.64 \\ 
                                         & ACN\cite{kd3}     & \underline{85.24}  & \underline{79.16} & \underline{78.65} & 86.72 & 85.87 & \underline{79.27} & 86.33 & 85.21 & 86.69 & 87.54 & 86.92 & 88.22 & 87.51 &  86.38 & 89.14 & 85.25$\pm$3.38, 20.43                                              \\ 
                                         & RFNet\cite{rfnet}     &   84.92 & 73.41 & 72.57 & \underline{87.93} & 86.22 & 78.31 & 89.43 & \textbf{86.81} & 89.98 & 89.23 & \underline{89.80} & \underline{90.22} & 90.16 & 86.95 & \underline{90.32}  & 85.75$\pm$6.03, 48.41                                   \\ 
                                         & mmFormer\cite{mmformer}     &   84.73 & 76.10 & 75.39 & \textbf{88.53} & \underline{86.75} & 79.24 & \underline{89.57} & 86.61 & \underline{90.05} & \underline{89.69} & 89.64 & 90.11 & \underline{90.20} & \underline{87.11} & 90.09  & \underline{86.25$\pm$5.16}, 62.41                          \\  
                                         & Ours     & \textbf{86.52} & \textbf{79.23} & \textbf{78.66} & 87.45 & \textbf{86.77} &  \textbf{79.60} & \textbf{89.94} &\underline{86.71} & \textbf{90.82} & \textbf{90.13} & \textbf{90.38} & \textbf{90.74} &\textbf{90.63} &  \textbf{88.09} & \textbf{91.26} & \textbf{87.12$\pm$4.43}                                    \\ 
\hline
\multirow{7}{*}{{TC}}                 &                     HeMIS\cite{hemis}     & 49.63  & 53.75 & 24.80 & 32.91 & 70.28 & 64.29 & 45.62 & 54.36 & 54.93 & 69.40 & 72.57 & 62.38 & 75.51 & 73.94 & 74.18 & 58.57$\pm$15.50, $0.01^\ast$                              \\ 
                                         & U-HVED\cite{hved}     & 56.62 & 64.50 & 36.77 & 54.38 & 74.46 & 65.29 & 59.03 & 58.66 & 62.57 & 73.14 & 75.85 & 63.72 & 73.52 & 76.81 & 72.96 & 64.55$\pm$10.72, $0.02^\ast$  \\ 
                                         & D2-Net\cite{d2net}     & 59.87 & 64.29 & 20.32 & 50.84 & 81.06 & 77.96 & 62.54 & 64.18 & 61.70 & 82.45 & 79.38 & 67.52 & 81.47 &  80.23 & 80.94 & 67.65$\pm$16.55, $2.02^\ast$
                                         \\ 
                                         & ACN\cite{kd3}     & 67.24 & \underline{84.35} & \underline{70.49} & 67.38 & \underline{84.70} & \underline{83.92} & 70.61 & \textbf{73.58} & 70.66 & 82.17 & \underline{84.35} & 67.08 & 81.94 & \underline{84.32} & \underline{84.73} & \underline{77.16$\pm$7.56}, 47.26                                                  \\ 
                                         & RFNet\cite{rfnet}     &  \underline{67.72} & 78.87 & 64.39 & \underline{67.85} & 83.04 & 80.84 & \underline{72.80} & 71.65 & \underline{73.32} & \underline{83.76} & 84.09 & \underline{74.89} & \underline{84.26} & 82.98 & 84.40  & 76.99$\pm$7.00, 41.76                                     \\ 
                                         & mmFormer\cite{mmformer}     &   65.92 & 77.50 & 62.94 & 66.10 & 80.58 & 79.35 & 72.31 & 69.89 & 71.39 & 79.72 & 81.53 & 73.30 & 80.68 & 80.56 & 81.62  & 74.89$\pm$6.46, 10.09                            \\  
                                         & Ours     & \textbf{68.12}  & \textbf{84.57} & \textbf{71.24} & \textbf{68.75} & \textbf{85.67} &  \textbf{84.39} & \textbf{73.48} & \underline{72.90} & \textbf{73.71} &  \textbf{84.97} & \textbf{85.43} & \textbf{75.62} & \textbf{84.75} &  \textbf{86.56} & \textbf{86.77} & \textbf{79.12$\pm$7.18}                                     \\
\hline
\multirow{7}{*}{{ET}}                 &                     HeMIS\cite{hemis}     & 22.47 & 56.20 & 7.89 & 9.64 & 64.07 & 65.66 & 17.73 & 26.95 & 27.42 & 65.83 & 70.35 & 30.18 & 68.97 & 69.52 & 73.80 & 45.11$\pm$24.97, $3.23^\ast$                               \\ 
                                         & U-HVED\cite{hved}     & 27.82  & 61.24 & 11.06 & 22.35  & 68.93 & 65.79 & 24.57 & 24.46 & 35.80 & 69.31 & 71.42 & 32.14 & 70.66 & 69.98 & 71.20 & 48.44$\pm$22.98, 6.41  
                                         \\ 
                                         & D2-Net\cite{d2net}     & 22.83 & 69.52 & 15.34 & 12.96 & 70.45 & 71.38 & 14.06 & 19.32 & 17.79 & 69.25 & 68.31 & 23.66 & 67.14 &  68.56 & 67.72 & 45.22$\pm$26.52, $4.07^\ast$
                                         \\ 
                                         & ACN\cite{kd3}     & \underline{43.26} & \textbf{78.57} & \underline{40.89} & \underline{42.14} & 74.95 & \underline{75.88} & 42.73 & \textbf{47.80} & 44.39 & \underline{76.72} & \underline{76.33} & 41.61 & 75.54 & \underline{75.27} & \underline{76.79} & \underline{60.85$\pm$17.12}, 78.65                                                 \\ 
                                         & RFNet\cite{rfnet}     & 40.62 & 69.73 & 37.62 & 38.08 & \underline{75.42} & 71.55 & \underline{45.67} & 43.44 & \underline{45.36} & 75.18 & \textbf{76.52} & \underline{47.14} & \underline{76.75} & 75.26 & 76.71  & 59.67$\pm$16.85, 64.25                                     \\ 
                                         & mmFormer\cite{mmformer}     &   39.65 & 66.23 & 37.08 & 38.72 & 68.70 & 67.84 & 45.15 & 42.61 & 43.69 & 68.42 & 68.36 & 45.33 & 68.45 & 69.81 & 68.16  & 55.88$\pm$13.86, 24.25                           \\  
                                         & Ours     & \textbf{44.87} & \underline{78.09} & \textbf{41.12} & \textbf{43.94} & \textbf{77.16} &  \textbf{77.58} & \textbf{45.81} & \underline{46.25} & \textbf{48.63} &  \textbf{77.29} & 76.04 & \textbf{48.22} & \textbf{77.92} &  \textbf{76.71} & \textbf{78.30} & \textbf{62.53$\pm$16.53}                                    \\
                        \hline
\end{tabular}
  \caption{Comparison with state-of-the-art (DSC $\%$) on segmentation of nested tumor regions (WT, TC, ET) for the different combinations of available modalities on BRATS2018. Our approach trains with 50\% full modality samples while others use 100\%. The best and second best scores are \textbf{bolded} and \underline{underlined}, respectively. Modalities present are denoted by $\bullet$, the missing ones by $\circ$. Statistically significant results with p-values $\le0.05$ are denoted by $\ast$.}
\label{tableseg}
\end{table*}

\section{Experiment Design and Results}
To validate our framework in various missing scenarios, we evaluate brain tumor
segmentation results on all fifteen combinations of the
four image modalities for a fixed test set. The average score is also reported
for comparisons.

\noindent\textbf{Datasets.}
We use three segmentation datasets from BRATS2018, BRATS2019, and BRATS2020 challenges~\cite{brats}. They comprise 285, 335, and 369 training cases respectively. All subjects have $M=4$ MR sequences. 
We perform 3D volumetric segmentation with images of size 155 $\times$ 240 $\times$ 240. 
Pre-processing details are contained in the supplementary (Sec. 15). The segmentation classes include whole tumor (WT), tumor core (TC), and enhancing tumor (ET). 
Additional segmentation results on two non-BRATS hetero-modal cohorts (with different medical imaging modalities such as CT and MRI) are also reported in the supplementary (Sec. 16).

\noindent\textbf{Implementation details.}
The experiments are implemented in Pytorch 1.7~\cite{pytorch} with three 48 GB Nvidia Quadro RTX 8000 GPUs. We drop modalities during training to construct the missing-modality dataset $\D^{m}$. We randomly sample a set of subjects from $\D^{m}_i$ assigned to each task distribution $P(\T^i)$. For a given subject, only those modalities which are not present in its associated task distribution are dropped. We ablate the fraction of subjects reserved for the full modality dataset $\D^{f}$(Fig.~\ref{figfull}). 
Our method adopts a modified Swin-UNETR~\cite{seg6} housing up to 4 encoders $E_1,E_2,E_3,E_4$ which are Swin transformers (see supplementary Sec. 12). 
Both MLPs for the discriminator and feature aggregation are fully connected networks whose hidden layer dimensions are 48 and 64 respectively. The images are first resized to 128 $\times$ 128 $\times$ 128, which is kept consistent across all compared methods. Features are extracted from 5 different levels of each encoder. For training and testing cohorts, we randomly split BRATS2018 into 200 and 85 subjects, BRATS2019 into 250 and 85 subjects, and BRATS2020 into 269 and 100 subjects, respectively. 
The batch size per task is kept as 1. During meta-training, we consider a metabatch size of 8, i.e., our meta-batch comprises 8 different modality combinations, each representing a separate task $\T_i$. 
More details can be found in the supplementary (Sec. 15). 
During inference, the meta-pretrained model is evaluated on test sets where all subjects have a fixed subset of modalities. The discriminator is discarded at inference. 

\noindent\textbf{Performance metrics.}
Dice similarity coefficient (DSC $\uparrow$) (Tab.~\ref{tableseg}) and Hausdorff Distance (HD95 $\downarrow$) (see supplementary Sec. 9) are used to evaluate segmentation performance. 

\subsection{Comparisons with State-of-the-art}
\noindent\textbf{Quantitative results:} In Tab.~\ref{tableseg}, we compare our approach with SOTA methods including HeMIS~\cite{hemis}, U-HVED~\cite{hved}, D2-Net~\cite{d2net}, ACN~\cite{kd3}, RFNet~\cite{rfnet} and mmFormer~\cite{mmformer} on BRATS2018. HeMIS, U-HVED, and D2-Net learn a biased mapping among available modalities and hence perform poorly compared to ACN which co-trains with the full modality of all samples. Recent shared latent space models like mmFormer and RFNet perform comparably to ACN. They either focus on learning inter-modal correlations or tumor region-aware fused representations. 
Our method utilizes full modality data from only 50\% samples, and yet outperforms these approaches. We thus excel in efficient utilization of full modality data. In comparison with the second-best approach in WT, TC, and ET, our average DSC shows improvements of 0.89\% (over mmFormer), 1.96\% (over ACN), and 1.68\% (over ACN), respectively. Although ACN pursues a KD-driven approach to achieve the partial-to-full modality mapping, it ends up building a combinatorial number of models dedicated to each subset. This leads to a highly ineffective solution which is also based on the impractical scenario that all training samples contain full modality data. In our framework, we mimic this distillation learning even in a shared latent model through efficient application of meta-learning and adversarial training. It can be seen from Tab.~\ref{tableseg} that our method surpasses all other approaches in 39 out of 45 multi-modal combinations across the three tumor regions despite being trained with only 50\% full modality samples. Our results are statistically significantly better (t-test, $p<0.05$) than HeMIS~\cite{hemis}, U-HVED~\cite{hved}, D2-Net~\cite{d2net}. Other methods (RFNet~\cite{rfnet}, mmFormer~\cite{mmformer}, ACN~\cite{kd3}) require full modality input for all samples, whereas ours does not.
However, for a fair comparison, we further trained 
~\cite{rfnet, mmformer, kd3} in a 50\% full modality setting, identical to ours (Tab.~\ref{tabfmablation}). It can be observed that our method outperforms the second-best approach by 11.78\%, 12.93\%, and 9.72\% in DSC on the WT, TC, and ET regions, respectively, on BRATS2018, clearly achieving SOTA performance. Evaluation via HD95 metric on BRATS2018 can be found in the supplementary (Sec. 9). 
Comparison with three SOTA methods on BRATS2020 are presented in Tab.~\ref{table2020}; Compared to the second-best approach, the average DSC of the three tumor areas is boosted by 1.78\%, 2.84\%, and 3.13\%, respectively. Detailed experiments on BRATS2020 and BRATS2019 have been provided in the supplementary (Sec. 10).

\begin{figure}[t]
  \includegraphics[width=1.0\linewidth]{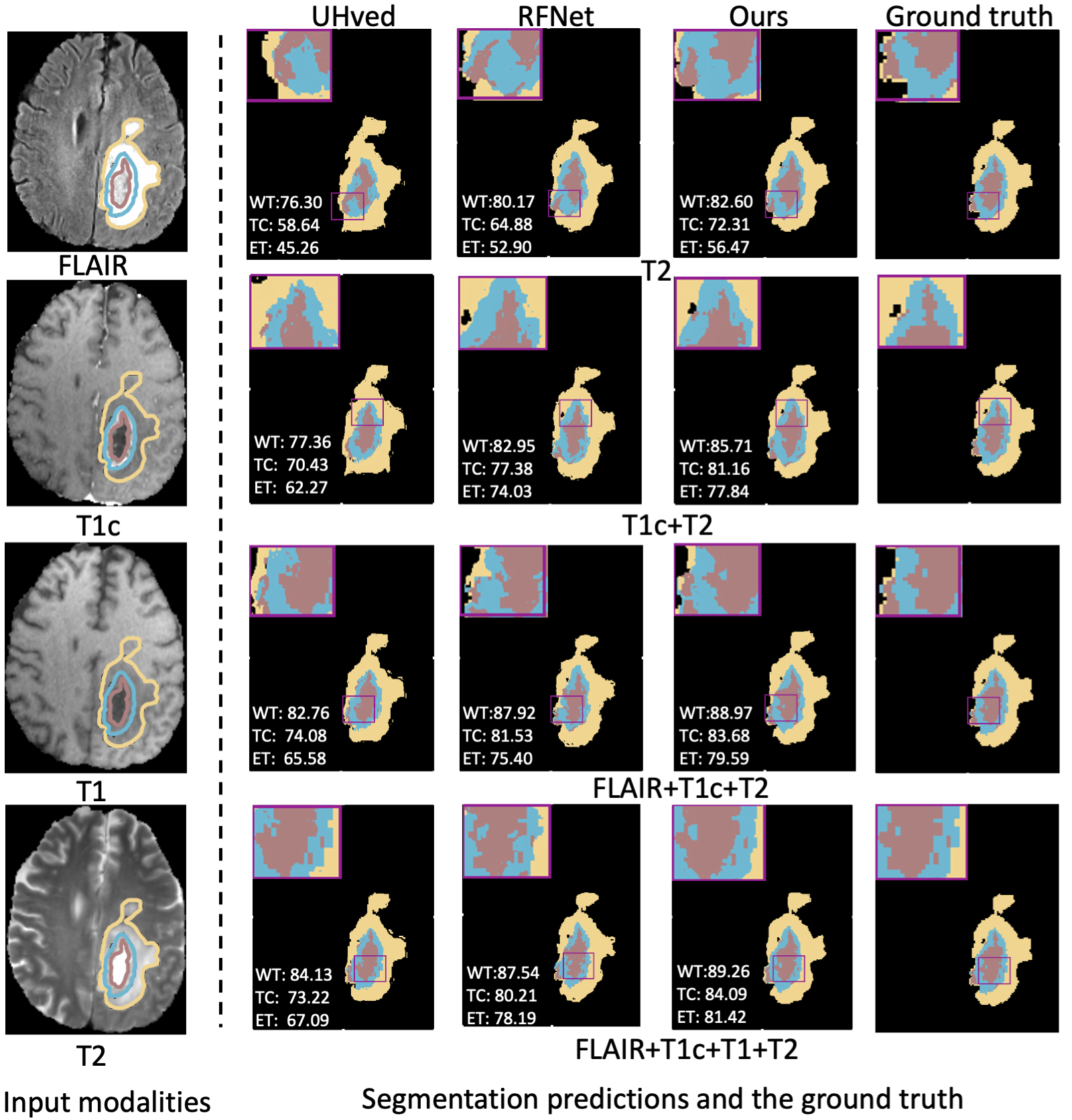}
    \caption{ Qualitative comparison. Column 1: four MRI modalities. Col 2-4: segmentation maps from three
methods for different combinations of modalities. Col 5: Ground truth. Our method is able to better capture gaps/islands (rows 1,3) and boundaries (row 4) in TC segmentations.}
    \label{segmap}
\end{figure}

\noindent\textbf{Qualitative results:} In Fig.~\ref{segmap} we visualize the segmentation masks predicted by U-HVED, RFNet, and ours from four combinations of modalities. Unlike others, our segmentations do not degrade sharply as additional modalities are dropped during the inference phase. Even with single T2 or T1c+T2 modalities, we achieve decent segmentation.
\subsection{Ablation Studies}
\begin{table*}[t]
  \begin{minipage}{0.32\textwidth}
  \footnotesize
    \centering
    \scriptsize
\begin{tabular}{c|ccc}
\hline
\multicolumn{1}{c|}{\multirow{2}{*}{Methods}} & \multicolumn{3}{c}{Average DSC (\%),  p-value $(10^{-2})$}                                       \\ \cline{2-4} 
\multicolumn{1}{c|}{}                         & \multicolumn{1}{c}{WT} & \multicolumn{1}{c}{TC} & ET \\ \hline
ACN\cite{kd3}                                       & \multicolumn{1}{c}{65.81, $0.01^\ast$}         &  \multicolumn{1}{l}{57.66, $0.01^\ast$}     &   47.36, 
$1.63^\ast$     \\
                                           RFNet\cite{rfnet}     &  \multicolumn{1}{c}{73.96, $0.01^\ast$}         & \multicolumn{1}{c}{62.37, $0.01^\ast$}     & 50.24, 
                                           $3.80^\ast$      \\
                                mmFormer\cite{mmformer}       &  \multicolumn{1}{c}{75.34, $0.01^\ast$}         & \multicolumn{1}{c}{66.19, $0.01^\ast$}   &  52.81, 
                                7.64        \\ 
                                Ours       &  \multicolumn{1}{c}{\textbf{87.12}}         & \multicolumn{1}{c}{\textbf{79.12}}     & \textbf{62.53}  
                                \\\hline
\end{tabular}
\caption{Comparison (DSC\%, p-value) on BRATS2018 with 50\% full modality}
\label{tabfmablation}
  \end{minipage}
  \hfill
  \begin{minipage}{0.32\textwidth}
  \footnotesize
    \centering
    \scriptsize
\begin{tabular}{c|ccc}
\hline
\multicolumn{1}{c|}{\multirow{2}{*}{Methods}} & \multicolumn{3}{c}{Average DSC (\%)}                                       \\ \cline{2-4} 
\multicolumn{1}{c|}{}                         & \multicolumn{1}{c}{WT} & \multicolumn{1}{c}{TC} & ET \\ \hline
mDrop                                       & \multicolumn{1}{c}{81.67}         &  \multicolumn{1}{l}{72.41}     &   52.06     \\ \hline
                                          + GAN     &  \multicolumn{1}{c}{84.49}         & \multicolumn{1}{c}{75.38}     & 55.64     \\
                                          + MetaL    &  \multicolumn{1}{c}{85.96}         & \multicolumn{1}{c}{77.23}     &  59.85      \\   
                                + GAN + MetaL       &  \multicolumn{1}{c}{\textbf{87.12}}         & \multicolumn{1}{c}{\textbf{79.12}}     &    \textbf{62.53}      \\ \hline
\end{tabular}
\caption{Ablation study demonstrating effectiveness of major components.}
\label{tableab1}
  \end{minipage}
  \hfill
 \begin{minipage}{0.32\textwidth}
  \footnotesize
		\centering
  \scriptsize
\begin{tabular}{c|ccc}
\hline
\multicolumn{1}{c|}{\multirow{2}{*}{Methods}} & \multicolumn{3}{c}{Average DSC (\%), p-value $(10^{-2})$}                                       \\ \cline{2-4} 
\multicolumn{1}{c|}{}                         & \multicolumn{1}{c}{WT} & \multicolumn{1}{c}{TC} & ET \\ \hline
HeMIS\cite{hemis}                                       & \multicolumn{1}{c}{77.29, $0.01^\ast$}         &  \multicolumn{1}{l}{67.12, $0.13^\ast$}     &   49.64, $4.25^\ast$     \\
                                           U-HVED\cite{hved}     &  \multicolumn{1}{c}{82.65, $1.49^\ast$}         & \multicolumn{1}{c}{69.08, $0.22^\ast$}     & 51.53, 5.84      \\
                                RFNet\cite{rfnet}       &  \multicolumn{1}{c}{86.96, 23.71}        & \multicolumn{1}{c}{78.79, 27.86}     &  62.14, 59.95        \\ 
                                Ours       &  \multicolumn{1}{c}{\textbf{88.74}}         & \multicolumn{1}{c}{\textbf{81.63}}     & \textbf{65.27}         \\\hline
\end{tabular}
\caption{Comparison (DSC\%, p-value) on BRATS2020.}
\label{table2020}

  \end{minipage}
\end{table*}

\noindent\textbf{Effectiveness of adversarial and meta-learning.}
We perform several ablations to evaluate and justify the contribution of each proposed module in our architecture. First, we remove both the Adversarial and the Meta-training strategies to perform segmentation from only fusion of available modalities. We thus formulate a baseline, \emph{mDrop}, where we include our feature-aggregation block to generate a fused representation from available modalities. \emph{mDrop} solely learns the intra-model relations through transformer encoders and inter-modal  dependencies through channel-weighted fusion.  The average DSC of our model outperforms \emph{mDrop} by 5.45\%, 6.71\%, and 10.47\% in the three tumor regions (Tab.~\ref{tableab1}). Hence it is evident that solely the modality-agnostic representations obtained from fusion of available modalities cannot generate accurate segmentations. This necessitates feature enrichment to improve the quality of the fused representation. 
We develop two variants through gradual introduction of our discriminator and meta-learning strategies as enrichment techniques. 
Both variants surpass mDrop considerably (Tab.~\ref{tableab1}). Meta-learning (MetaL) proved to be better since we built the heterogeneous task distribution with modality combinations (to reduce bias) and also explicitly adapted to the full modality feature space efficiently. Finally, we arrive at an end-to-end meta-learning framework that also benefits from auxiliary supervision provided by the adversarial discriminator. 

\begin{figure}[hbtp]
\centering
    \begin{subfigure}{0.42\linewidth}
  \includegraphics[width=1.0\linewidth]{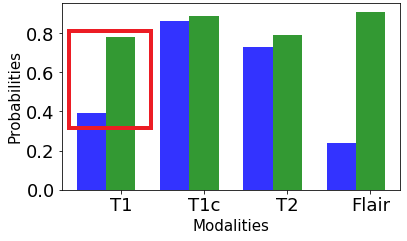}
    \caption{\emph{mDrop}}
    \label{base}
  \end{subfigure}
  \begin{subfigure}{0.42\linewidth}
  \includegraphics[width=1.0\linewidth]{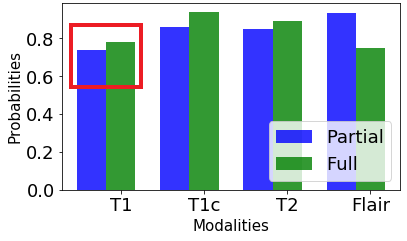}
    \caption{Ours}
    \label{enhance}
  \end{subfigure}
  \caption{Comparison of baseline and enhanced features}
  \label{compare}
\end{figure}

\noindent\textbf{Evaluation of enhanced representations.}
\label{sec:enhanceexpt}
To evaluate the quality of enhanced representations, we designed a simple experiment. We first extract the bottleneck fused representations $\mathbf{F}^{B}_{fused}$ of 50 test subjects for both scenarios of full modality (where all are present) and partial modality (where only T1c, T2 are present). This was done for both \emph{mDrop} as well as our approach. The fused representations were fed into a classifier trained to predict the probabilities of a modality's presence. The average probabilities obtained from our method attain comparable distributions across partial and full modality settings (Fig.~\ref{compare}), depicting the desired enhancement of  $\mathbf{F}^{B}_{fused}$. However, for \emph{mDrop}, probabilities of T1c and T2 being present are much higher than T1, and FLAIR in the missing scenario. Hence the relevant information from the latter two modalities is being lost. The red boxes depicts how our probability for predicting T1 is considerably higher than \emph{mDrop} even in T1-missing scenario.


\begin{figure}[hbtp]
    \begin{subfigure}{0.49\linewidth}
  \includegraphics[width=0.9\linewidth]{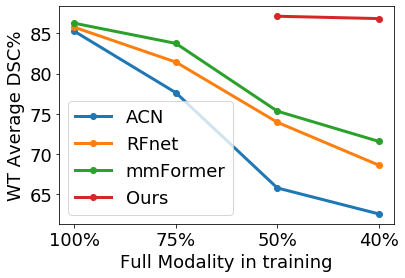}
    \caption{Ablation results for varying \% of full modality in training.}
    \label{figfull}
  \end{subfigure}
  \begin{subfigure}{0.49\linewidth}
  \includegraphics[width=0.9\linewidth]{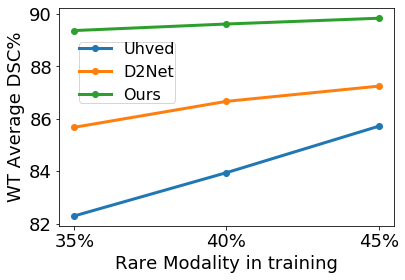}
    \caption{Ablation results for varying \% of FLAIR in training.}
    \label{figrare}
  \end{subfigure}
  \caption{Ablation studies.}
\end{figure}

\noindent\textbf{Robustness to full modality setting.}
Due to the meta-learning strategy incorporated while training on hybrid data, we hypothesize that our network is robust to the ratio of full modality samples used in training. We compare against ACN, RFNet, and mmFormer by varying the full modality count from 100\% to 40\% (Fig.~\ref{figfull}). In order to retain sufficient samples for each combination task in meta-training, we assume that at least 50\% of the subjects have partial modalities. Hence we show our results only on 50\% and 40\% proportions of full modality data. The fact that even with 50\% full modality samples, we match the evaluation scores of SOTA at 100\% setting is noteworthy. A sharp degradation can be noticed in the average WT DSC of SOTA once the number of full modality data decreases. On the other hand, our method shows only a minor drop of 0.29\%. This is due to ACN being heavily dependent on the full modality for knowledge distillation. RFNet and mmFormer require full modality data as input to the network. They under-fit since their overall sample count decreases. Our method efficiently utilizes even limited samples of full modality data for feature adaptation in meta-testing. 
Owing to the above reasons, our approach is resilient to change in full modality proportion. Results for other tumor regions are provided in supplementary (Sec. 8).

\noindent\textbf{Bias to presence of a specific modality.}
Our model is robust to the scenario when a modality appears rarely during training. Tab.~\ref{tableabrare} demonstrates that when only 35\% of FLAIR is considered for training, our method consistently outperforms U-HVED and D2-Net in all the 8 inference scenarios involving FLAIR. Our average DSC (89.37, 78.11, and 60.80) for (WT, TC, and ET) are significantly higher than the second-best method (85.68, 68.77, and 47.07). We attribute this improvement to meta-learning which precludes the model from learning a biased mapping among available modalities by aligning the shared representations to full modality representations. Further experiments in Fig.~\ref{figrare} demonstrate that the performance of U-HVED and D2-Net are highly sensitive to the availability of a particular modality while our approach is impervious to this. On increasing FLAIR from 35\% to 45\%, our WT gain (+0.47\%) is much lower than U-HVED (+3.42\%) or D2-Net (+1.57\%). Experiments with another modality (T1c) are provided in supplementary (Sec. 11).

\noindent\textbf{Robustness to backbone variants.}
The proposed meta and adversarial training strategies are robust to any employed backbone including 3DUnet~\cite{3dunet}, nnUnet~\cite{nnunet} and AttentionUnet~\cite{attunet}. Comparisons are provided in supplementary (Sec. 7).

\noindent\textbf{Ablation on aggregation block.}
We design 3 baseline aggregation
modules to highlight the effectiveness of our fusion strategy. Architectural details are provided in supplementary (Sec. 13).

\setlength{\tabcolsep}{2pt}
\begin{table}[t]
\fontsize{7}{9}\selectfont
		\centering
\begin{tabular}{c|c|c|c|c|c|c|c|c|c|c}
\hline
\multicolumn{1}{c|}{\multirow{4}{*}{M}} & FLAIR & $\bullet$ & $\bullet$ & $\bullet$ & $\bullet$ &$\bullet$  & $\bullet$ & $\bullet$ & $\bullet$ & \multirow{4}{*}{Avg} 
\\ 
\multicolumn{1}{c|}{}                   & T1    &  $\circ$ & $\bullet$ &  $\circ$ & $\circ$ & $\bullet$ & $\bullet$ & $\circ$  & $\bullet$ &                            \\ 
\multicolumn{1}{c|}{}                   & T1c  & $\circ$  & $\circ$ & $\circ$ & $\bullet$ & $\bullet$ & $\circ$ & $\bullet$ & $\bullet$ &                             \\
\multicolumn{1}{c|}{}                   & T2    &  $\circ$ & $\circ$ & $\bullet$  & $\circ$ & $\circ$ & $\bullet$ & $\bullet$ & $\bullet$ &                            \\ \hline
\multirow{4}{*}{{WT}}                 &                     U-HVED    &  76.61 & 78.94 & 83.98 & 80.47 & 82.76 & 84.39 & 85.08 & 86.25 & 82.30                             \\ 
                                         & D2-Net     & 82.44  & 83.79 & 85.16 & 85.55 & 85.18 & 87.63 & 88.02 & 87.70 &  \underline{85.68}                                                                \\ 
                                         & Ours     &  \textbf{86.53} &  \textbf{88.77} & \textbf{90.05} & \textbf{89.41} & \textbf{89.49} & \textbf{90.13} & \textbf{90.07} & \textbf{90.55} &\textbf{89.37}                                    \\ 
\hline
\multirow{4}{*}{{TC}}                 &                     U-HVED     &  50.23 & 55.46 & 57.35 & 71.85 & 74.31 & 59.02 & 72.19 & 71.40 & 63.97                              \\ 
                                         & D2-Net    & 47.19  & 59.98 & 59.65 & 81.23 & 77.44 & 64.86 & 80.11 & 79.70 &  \underline{68.77}                                                                \\ 
                                         & Ours     &  \textbf{67.26} &  \textbf{72.28} & \textbf{72.64} & \textbf{84.03} & \textbf{84.87} & \textbf{74.08} & \textbf{83.86} & \textbf{85.91} & \textbf{78.11}                                    \\ 
\hline
\multirow{4}{*}{{ET}}                 &                     U-HVED     &  17.58 & 20.39 & 32.17& 67.75 & 70.11 & 28.94 & 69.60 & 70.09 & \underline{47.07}                              \\ 
                                         & D2-Net     &  9.37 & 11.78 & 14.09 & 68.14 & 67.58 & 20.22 & 66.31 & 66.80 & 40.53                                                                  \\ 
                                         & Ours     & \textbf{42.58}  &  \textbf{44.27} & \textbf{47.14} &\textbf{76.12} & \textbf{74.95} & \textbf{46.83} & \textbf{76.91} & \textbf{77.61} & \textbf{60.80}                                    \\ 
\hline
\end{tabular}
\caption{Results for rare occurrence of FLAIR in training.}
\label{tableabrare}

\end{table}

\section{Conclusion}
We present a novel training strategy to address the problem of missing modalities in brain tumor segmentation under limited full modality supervision. We adopt meta-learning and formulate modality combinations as separate meta-tasks to mitigate the bias towards modalities rarely encountered in training. We distill discriminative features from full modality data in the meta-testing phase, thereby discarding the impractical omnipresence of full modalities for all samples. This mapping is further co-supervised by novel adversarial learning in latent space, that guarantees the generation of superior modality-agnostic representations.
In the future we will validate our method on other downstream tasks such as radiogenomics classification~\cite{singh2021radiomics} and treatment response prediction~\cite{prasanna2017radiomic}. 



\section{Acknowledgements}
Reported research was partially supported by NIH 1R21CA258493-01A1 and NSF CCF-2144901. The content is solely the responsibility of the authors and does not necessarily represent the official views of the National Institutes of Health.


{\small
\bibliographystyle{ieee_fullname}
\bibliography{egbib}
}

\section*{Supplementary Material}
In the supplementary material, we provide additional information to better understand the contributions and claims of our proposed work. 
First, the ablation results for various encoder-decoder backbones (3DUnet, nnUnet, AttentionUnet) are shown in Sec.~\ref{backbonesupp}. Our method also attains state-of-the-art performance when re-implemented with a 3DUnet backbone (like others). Note that we always maintain 50\% full modality as the default setup for our approach. Other methods are however re-implemented with two different proportions (100\% and 50\% full modality data). In Sec.~\ref{fullsupp} we demonstrate the robustness of our approach to varying proportions of full modality data. Ablation results are provided for additional tumor regions. Evaluations via an additional metric, Hausdorff Distance, on BRATS2018 and BRATS2020 are shown in Sec.~\ref{metric}. Comprehensive results on BRATS2019 and BRATS2020 datasets are provided in Sec.~\ref{brats2020supp}. In Sec.~\ref{raresupp}, further experiments are conducted to test the bias of our model to the occurrence of a specific modality (FLAIR or T1c) as input during training. In Sec.~\ref{fusioneq} we discuss the fusion strategy in more detail through equations. Architectural details and experiments with different fusion baselines are demonstrated in Sec.~\ref{fusesupp}. Additional qualitative segmentation maps are shown in Sec.~\ref{segsupp}. Further details regarding the implementation, including pre-processing steps, are outlined in Section~\ref{implement}. The segmentation performance of our model on additional non-BRATS datasets can be found in Sec.~\ref{nonbrats}.

\section{Ablation results on robustness to encoder-decoder backbones}
\label{backbonesupp}

Our proposed meta-learning and adversarial training strategies are independent of the backbones utilized in the framework. We evaluate our approach using different backbones including 3DUnet~\cite{3dunet}, nnUnet~\cite{nnunet}, and AttentionUnet~\cite{attunet}. The average DSCs reported in Tab.~\ref{tableenc} vary marginally between 1.25\% and 2.6\% across all encoder-decoder variants, highlighting the backbone-agnostic nature of our framework. 
A schematic of the adopted Swin-UNETR encoder is provided in Fig.~\ref{swin}.

\setlength{\tabcolsep}{2pt}
\begin{table}[!hbtp]
\centering
\begin{tabular}{c|ccc}
\hline
\multicolumn{1}{c|}{\multirow{2}{*}{Methods}} & \multicolumn{3}{c}{Average DSC(\%),  p-value $(10^{-2})$}                                       \\ \cline{2-4} 
\multicolumn{1}{c|}{}                         & \multicolumn{1}{c}{WT} & \multicolumn{1}{c}{TC} & ET \\ \hline
3DUnet~\cite{3dunet}                                       & \multicolumn{1}{c}{85.70, 42.04}         &  \multicolumn{1}{l}{77.87, 67.28}     &   59.93, 67.41     \\
                                           AttentionUnet~\cite{attunet}     &  \multicolumn{1}{c}{86.02, 52.05}         & \multicolumn{1}{c}{78.05, 71.10}     &  60.46, 73.13    \\
                                            nnUnet~\cite{nnunet}    &  \multicolumn{1}{c}{86.53, 72.47}         & \multicolumn{1}{c}{78.64, 86.16}     & 62.28, 96.69       \\   
                                Ours       &  \multicolumn{1}{c}{\textbf{87.12}}         & \multicolumn{1}{c}{\textbf{79.12}}     & \textbf{62.53}         \\\hline
\end{tabular}
\caption{Ablation on backbone variants}
\label{tableenc}
\end{table}

For the convenience of comparison, we have listed all the model performances (implemented using 3D-Unet backbone) in Tab~\ref{table3d}. It should be noted that even with 3D-Unet as the backbone, our proposed method achieves results comparable to SOTA. This performance improvement may be attributed to the proposed meta and adversarial learning techniques, rather than the choice of backbone. However, our framework is trained with only 50\% full modality samples, unlike other approaches that utilize full modality for all patients (100\%).  

\setlength{\tabcolsep}{2.5pt}
\begin{table}[hbtp]
\centering
\scriptsize
\begin{tabular}{c|ccc}
\hline
\multicolumn{1}{c|}{\multirow{2}{*}{Methods}} & \multicolumn{3}{c}{Average DSC (\%)}                                     \\ \cline{2-4} 
\multicolumn{1}{c|}{}                         & \multicolumn{1}{c}{WT} & \multicolumn{1}{c}{TC} & ET \\ \hline
HeMIS~\cite{hemis}                                     & \multicolumn{1}{c}{76.23}         &  \multicolumn{1}{l}{58.57}     &   45.11    \\
                                           U-HVED~\cite{hved}     &  \multicolumn{1}{c}{79.55}         & \multicolumn{1}{c}{64.55}     & 48.44      \\
                                D2-Net~\cite{d2net}       &  \multicolumn{1}{c}{77.04}         & \multicolumn{1}{c}{67.65}     &  45.22        \\
                                ACN~\cite{kd3}       &  \multicolumn{1}{c}{85.25}         & \multicolumn{1}{c}{77.16}     &  {\textbf{60.85}}        \\
                                RFNet~\cite{rfnet}       &  \multicolumn{1}{c}{85.75}         & \multicolumn{1}{c}{76.99}     &  59.67        \\
                                mmFormer~\cite{mmformer}       &  \multicolumn{1}{c}{\textbf{86.25}}         & \multicolumn{1}{c}{74.89}     &  55.88        \\
                                Ours (3D-Unet)      &  \multicolumn{1}{c}{85.70}         & \multicolumn{1}{c}{\textbf{77.87}}    & 59.93       \\\hline
\end{tabular}
\caption{Comparison on BRATS2018 with 3D-Unet backbone. All methods here are implemented with 3D-Unet. Only our approach is trained with 50\% full modality samples while HeMIS, U-HVED. D2-Net, ACN, RFNet, and mmFormer are trained with 100\% full modality samples.}
\label{table3d}
\end{table}

Moreover, methods like mmFormer, RFNet, and ACN \textit{always require full-modality data} as input. For a fair comparison, we have demonstrated in Tab.~\ref{tab50} that, if considering only 50\% full-modality data as input (like ours), there is a significant drop in performance for all other methods.

\setlength{\tabcolsep}{2.5pt}
\begin{table}[hbtp]
\centering
\begin{tabular}{c|ccc}
\hline
\multicolumn{1}{c|}{\multirow{2}{*}{Methods}} & \multicolumn{3}{c}{Average DSC (\%),  p-value $(10^{-2})$}                                       \\ \cline{2-4} 
\multicolumn{1}{c|}{}                         & \multicolumn{1}{c}{WT} & \multicolumn{1}{c}{TC} & ET \\ \hline
ACN\cite{kd3}                                       & \multicolumn{1}{c}{65.81, $0.01^\ast$}         &  \multicolumn{1}{l}{57.66, $0.01^\ast$}     &   47.36, 
$1.50^\ast$     \\
                                           RFNet\cite{rfnet}     &  \multicolumn{1}{c}{73.96, $0.01^\ast$}         & \multicolumn{1}{c}{62.37, $0.01^\ast$}     & 50.24, 
                                           $3.95^\ast$      \\
                                mmFormer\cite{mmformer}       &  \multicolumn{1}{c}{75.34, $0.01^\ast$}         & \multicolumn{1}{c}{66.19, $0.01^\ast$}   &  52.81, 
                                8.79        \\ 
                                Ours (3D-Unet)      &  \multicolumn{1}{c}{\textbf{85.70}}         & \multicolumn{1}{c}{\textbf{77.87}}     & \textbf{59.93} 
                                \\ \hline
\end{tabular}
\caption{Comparison (DSC\%, p-value) on BRATS2018 with 3D-Unet backbone. All methods here are implemented with 3D-Unet, and trained with 50\% full modality samples.}
\label{tab50}
\end{table}

\section{Additional ablation results on robustness to full modality}
\label{fullsupp}
\begin{figure}[t]
    \begin{subfigure}{0.49\linewidth}
  \includegraphics[width=1\linewidth]{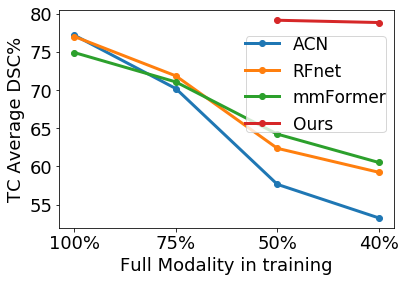}
    \label{figfull1}
  \end{subfigure}
  \begin{subfigure}{0.49\linewidth}
  \includegraphics[width=1\linewidth]{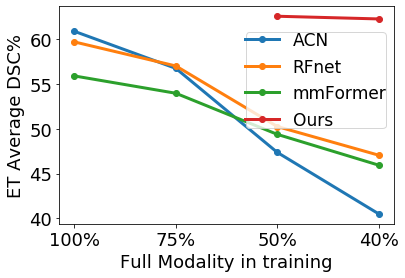}
    \label{figfull2}
  \end{subfigure}
  \caption{Ablation studies for varying \% of full modality in training.}
  \label{figfullsup}
\end{figure}

Ablation results on the WT region have been provided in the main paper (Sec.~4.2, Fig.~7a) to demonstrate that our method performs well even with a limited number of full modality samples in training. Here we are providing additional results for TC and ET regions. We compare against ACN~\cite{kd3}, RFNet~\cite{rfnet}, and mmFormer~\cite{mmformer} by varying the full modality count from 100\% to 40\% (Fig.~\ref{figfullsup}). In order to retain sufficient samples for each combination task in meta-training, we assume that at least 50\% of the patients have partial modalities. Hence
we show our results only on 50\% and 40\% proportions of
full modality data. Unlike other methods, ours shows only a minor decline in DSC (0.3\%) for both TC and ET. These experimental results further support the claim 
that our proposed method is robust to full modality setting.

We also present the results (Tab.~\ref{scarce}) achieved by our method when trained with 10\% and 20\% full modality samples. Notably, our method still generates high dice scores even in such severely missing modality scenarios. Please note that it was not possible to train SOTA methods in this scenario since they are left with only $\approx$20 or $\approx$40 subjects. 

\begin{table}[ht]
\centering
\begin{tabular}{c|ccc}
\hline
\multicolumn{1}{c|}{\multirow{2}{*}{Settings}} & \multicolumn{3}{c}{Average DSC (\%)}                                       \\ \cline{2-4} 
\multicolumn{1}{c|}{}                         & \multicolumn{1}{c}{WT} & \multicolumn{1}{c}{TC} & ET \\ \hline
10\% FM                                       & \multicolumn{1}{c}{81.56}         &  \multicolumn{1}{l}{72.89}     &   56.70     \\
                                          20\% FM     &  \multicolumn{1}{c}{84.41}         & \multicolumn{1}{c}{76.63}     & 60.82
                                          \\
                                          50\% FM     &  \multicolumn{1}{c}{87.12}         & \multicolumn{1}{c}{79.12}     & 62.53 \\\hline
\end{tabular}
\hfill
\caption{Ablation on BRATS2018 when trained with an extremely low proportion of full modality samples.}
\label{scarce}
\end{table}

\section{Additional metric (Hausdorff distance)}
\label{metric}
Model evaluations have also been performed using Hausdorff Distance (HD95) on BRATS2018 and BRATS2020, respectively. The results can be found in Tab.~\ref{stdpvhd} and ~\ref{stdpvhd20}. It can be observed from Tab.~\ref{stdpvhd} that our method significantly outperforms SOTA in 2/3 tumor regions (WT, TC) and emerges second-best for ET on BRATS2018; noting that all other methods are trained with 100\% full modality samples, ours is only 50\%.

\setlength{\tabcolsep}{2pt}
\begin{table}[hbtp]
\centering
\scriptsize
\begin{tabular}{c|ccc}
\hline
\multicolumn{1}{c|}{\multirow{2}{*}{Methods}} & \multicolumn{3}{c}{Average HD95 $(\downarrow)$, p-value $(10^{-2})$}               \\ \cline{2-4} 
\multicolumn{1}{c|}{}                         & \multicolumn{1}{c}{WT} & \multicolumn{1}{c}{TC} & ET \\ \hline
HeMIS~\cite{hemis}                                      & \multicolumn{1}{c}{14.85$\pm$7.32, $0.08^\ast$}         &  \multicolumn{1}{l}{15.58$\pm$8.44, $0.16^\ast$}     &   19.65$\pm$12.37, $0.25^\ast$     \\
                                           U-HVED~\cite{hved}     &  \multicolumn{1}{c}{13.64$\pm$6.27, $0.12^\ast$}         & \multicolumn{1}{c}{14.91$\pm$7.19, $0.09^\ast$}     & 18.43$\pm$11.68, $0.42^\ast$      \\
                                D2-Net~\cite{d2net}       &  \multicolumn{1}{c}{10.82$\pm$6.70, 7.75}         & \multicolumn{1}{c}{11.76$\pm$7.35, 5.12}     &  14.79$\pm$8.79, $1.75^\ast$        \\
                                ACN~\cite{kd3}       &  \multicolumn{1}{c}{8.15$\pm$2.03, 39.05}         & \multicolumn{1}{c}{9.37$\pm$2.81, 8.39}     &  \textbf{8.62$\pm$2.43}, 86.77      \\
                                RFNet~\cite{rfnet}       &  \multicolumn{1}{c}{7.89$\pm$1.72, 58.64}         & \multicolumn{1}{c}{8.43$\pm$2.52, 42.09}     &  12.56$\pm$3.68, $0.36^\ast$        \\
                                mmFormer~\cite{mmformer}      &  \multicolumn{1}{c}{\underline{7.67$\pm$2.14}, 84.97}         & \multicolumn{1}{c}{\underline{8.06$\pm$2.41}, 69.59}     &  10.54$\pm$3.13, 11.41        \\
                                Ours       &  \multicolumn{1}{c}{\textbf{7.53$\pm$1.86}}         & \multicolumn{1}{c}{\textbf{7.73$\pm$2.16}}     & \underline{8.78$\pm$2.77}         \\\hline
\end{tabular}
\caption{Comparison on BRATS2018 with HD95. The best and second best scores are \textbf{bolded} and \underline{underlined}, respectively.}
\label{stdpvhd}
\end{table}

\begin{table}[hbtp]
\centering
\scriptsize
\begin{tabular}{c|ccc}
\hline
\multicolumn{1}{c|}{\multirow{2}{*}{Methods}} & \multicolumn{3}{c}{Average HD95 $(\downarrow)$, p-value $(10^{-2})$}               \\ \cline{2-4} 
\multicolumn{1}{c|}{}                         & \multicolumn{1}{c}{WT} & \multicolumn{1}{c}{TC} & ET \\ \hline
HeMIS~\cite{hemis}                                      & \multicolumn{1}{c}{14.41$\pm$7.14, $0.11^\ast$}         &  \multicolumn{1}{l}{15.13$\pm$8.29, $0.21^\ast$}     &   19.24$\pm$12.07, $0.26^\ast$     \\
                                           U-HVED~\cite{hved}     &  \multicolumn{1}{c}{13.32$\pm$6.11, $0.13^\ast$}         & \multicolumn{1}{c}{14.74$\pm$6.97, $0.08^\ast$}     & 18.26$\pm$11.53, $0.41^\ast$      \\
                               
                                RFNet~\cite{rfnet}       &  \multicolumn{1}{c}{\underline{7.66$\pm$1.74}, 76.06}         & \multicolumn{1}{c}{\underline{8.27$\pm$2.45}, 44.00}     &  \underline{12.38$\pm$3.72}, $0.44^\ast$        \\
                                Ours       &  \multicolumn{1}{c}{\textbf{7.46$\pm$1.82}}         & \multicolumn{1}{c}{\textbf{7.60$\pm$2.23}}     & \textbf{8.69$\pm$2.72}        \\\hline
\end{tabular}
\caption{Comparison on BRATS2020 with HD95. The best and second best scores are \textbf{bolded} and \underline{underlined}, respectively.}
\label{stdpvhd20}
\end{table}

\setlength{\tabcolsep}{4pt}
\begin{table*}[hbtp]
\footnotesize
\begin{tabular}{c|c|c|c|c|c|c|c|c|c|c|c|c|c|c|c|c|c}
\hline
\multicolumn{1}{c|}{\multirow{4}{*}{M}} & FLAIR & $\circ$ & $\circ$ & $\circ$ & $\bullet$ & $\circ$  & $\circ$  & $\bullet$ & $\circ$ & $\bullet$ & $\bullet$ &$\bullet$  & $\bullet$ & $\bullet$ & $\circ$ & $\bullet$ & \multirow{4}{*}{Avg} 
\\ 
\multicolumn{1}{c|}{}                   & T1    &  $\circ$ & $\circ$  &$\bullet$  & $\circ$ & $\circ$ & $\bullet$ & $\bullet$ & $\bullet$ & $\circ$ & $\circ$ & $\bullet$ & $\bullet$ & $\circ$  & $\bullet$ & $\bullet$ &                            \\ 
\multicolumn{1}{c|}{}                   & T1c  & $\circ$ &$\bullet$ & $\circ$ & $\circ$ &$\bullet$  &$\bullet$  & $\circ$ & $\circ$ & $\circ$ & $\bullet$ & $\bullet$ & $\circ$ & $\bullet$ & $\bullet$ & $\bullet$ &                             \\
\multicolumn{1}{c|}{}                   & T2    & $\bullet$ & $\circ$ & $\circ$ & $\circ$ & $\bullet$ & $\circ$ & $\circ$ &$\bullet$  &$\bullet$  & $\circ$ & $\circ$ & $\bullet$ & $\bullet$ & $\bullet$ & $\bullet$ &                            \\ \hline
\multirow{4}{*}{{WT}}                 &                     HeMIS\cite{hemis}     & 80.34 & 66.92 & 66.35 & 58.72 & 85.16 & 73.41 & 69.79 & 83.30 & 83.76 & 73.41 & 76.78 & 84.43 & 85.17 & 85.84 & 86.03 & 77.29                           \\ 
                                         & U-HVED\cite{hved}     & 82.13 & 71.42 & 58.30 & 82.76 & 85.72 & 74.09 & 86.46 & 84.34 & 87.91 & 87.15 & 86.59 & 88.66 & 88.92 & 85.86 & 89.43 & 82.65
                                          \\ 
                                         & RFNet\cite{rfnet}     & 86.30 & 76.34 & 77.72 & 87.05 & 88.02 & 81.07 & 89.72 & 88.02 & 89.64 & 89.51 & 90.44 & 90.62 & 90.55 & 88.50 & 91.01 & \underline{86.96}                                 \\ 
                                         & Ours     & \textbf{88.24} & \textbf{82.29} & \textbf{83.41} & \textbf{88.37} & \textbf{88.78} & \textbf{83.26} & \textbf{90.52} & \textbf{89.66} & \textbf{90.55} & \textbf{90.83} & \textbf{91.34} & \textbf{91.68} & \textbf{91.17} & \textbf{89.49} & \textbf{91.57} & \textbf{88.74}                                   \\ 
\hline
\multirow{4}{*}{{TC}}                 &                     HeMIS\cite{hemis}     & 60.83 & 74.22 & 48.57 & 37.03 & 79.84 & 78.35 & 48.19 & 60.80 & 60.21 & 74.62 & 78.88 & 63.48 & 79.24 & 81.56 & 81.03 & 67.12                           \\ 
                                         & U-HVED\cite{hved}     & 61.37 & 74.93 & 39.54 & 52.42 & 80.27 & 79.11 & 57.38 & 62.17 & 63.47 & 77.45 & 79.02 & 65.39 & 80.19 & 81.72 & 81.68 & 69.07   \\ 
                                         & RFNet\cite{rfnet}     & 70.94 & 82.45 & 65.58 & 69.88 & 85.82 & 83.88 & 72.76 & 72.90 & 73.45 & 85.71 & 85.97 & 74.74 & 86.11 & 85.55 & 86.24 & \underline{78.79}                                    \\ 
                                         & Ours     & \textbf{73.56} & \textbf{86.37} & \textbf{74.69} & \textbf{72.33} & \textbf{87.71} & \textbf{87.52} & \textbf{75.94} & \textbf{74.50} & \textbf{76.24} & \textbf{87.79} & \textbf{87.82} & \textbf{76.93} & \textbf{87.31} & \textbf{87.98} & \textbf{87.75} & \textbf{81.63}                                    \\
\hline
\multirow{4}{*}{{ET}}                 &                     HeMIS\cite{hemis}     & 32.78 & 64.95 & 20.41 & 14.63 & 71.12 & 71.40 & 19.04 & 29.76 & 30.66 & 69.52 & 71.39 & 32.13 & 71.98 & 72.37 & 72.44 & 49.64                             \\ 
                                         & U-HVED\cite{hved}     & 31.86 & 68.43 & 18.21 & 25.85 & 70.48 & 70.79 & 27.94 & 32.37 & 33.64 & 71.24 & 72.16 & 34.48 & 71.72 & 71.92 & 71.87 & 51.53  
                                           \\ 
                                         & RFNet\cite{rfnet}     & 48.03 & 74.84 & 36.58 & 38.45 & 76.66 & 76.52 & 43.12 & 51.40 & 51.02 & 76.38 & 77.10 & 49.82 & 77.07 & 78.10 & 77.02 & \underline{62.14}                                  \\  
                                         & Ours     & \textbf{52.77} & \textbf{80.06} & \textbf{42.28} & \textbf{44.87} & \textbf{78.92} & \textbf{79.85} & \textbf{46.73} & \textbf{54.67} & \textbf{54.29} & \textbf{78.81} & \textbf{77.31} & \textbf{50.69} & \textbf{79.24} & \textbf{79.43} & \textbf{79.12} & \textbf{65.27}                                   \\
                        \hline
\end{tabular}
  \caption{Comparison with state-of-the-art for the different combinations of available modalities on BRATS2020. Dice scores (DSC $\%$) are computed for three nested tumor subregions - Whole tumor (WT), Tumor core (TC), Enhancing tumor (ET). Modalities present are denoted by $\bullet$, the missing ones by $\circ$. The best and second best scores are \textbf{bolded} and \underline{underlined}, respectively.}
\label{tablesegsupp}
\end{table*}

\section{Results on BRATS2019 and BRATS2020 datasets}
\label{brats2020supp}
In Tab.~\ref{tablesegsupp}, we compare our approach
with three state-of-the-art methods including HeMIS~\cite{hemis}, U-HVED~\cite{hved}, and RFNet~\cite{rfnet} for tumor segmentation on BRATS2020 dataset~\cite{brats}. The average DSCs of the three tumor areas are boosted by 1.78\%, 2.84\%, and 3.13\%, respectively. A similar comparison on the BRATS2019 dataset is shown in Tab.~\ref{tab19}, where the average DSC scores are boosted by 1.57\%, 2.83\%, and 2.94\%.

\setlength{\tabcolsep}{2pt}
\begin{table}[hbtp]
\centering
\scriptsize
\begin{tabular}{c|ccc}
\hline
\multicolumn{1}{c|}{\multirow{2}{*}{Methods}} & \multicolumn{3}{c}{Average DSC (\%)}                                     \\ \cline{2-4} 
\multicolumn{1}{c|}{}                         & \multicolumn{1}{c}{WT} & \multicolumn{1}{c}{TC} & ET \\ \hline
HeMIS~\cite{hemis}                                     & \multicolumn{1}{c}{76.69}         &  \multicolumn{1}{l}{64.37}     &   48.24    \\
                                           U-HVED~\cite{hved}     &  \multicolumn{1}{c}{81.53}         & \multicolumn{1}{c}{67.81}     & 50.25      \\
                                RFNet~\cite{rfnet}       &  \multicolumn{1}{c}{86.49}         & \multicolumn{1}{c}{77.92}     &  60.88        \\
                                Ours      &  \multicolumn{1}{c}{\textbf{88.06}}       & \multicolumn{1}{c}{\textbf{80.75}}    & \textbf{63.82}       \\\hline
\end{tabular}
\caption{Comparison (DSC $\%$) on BRATS2019}
\label{tab19}
\end{table}

\section{Additional ablation results on bias to presence of a specific modality}
\label{raresupp}
\begin{figure}[hbtp]
    \begin{subfigure}{0.49\linewidth}
  \includegraphics[width=1.0\linewidth]{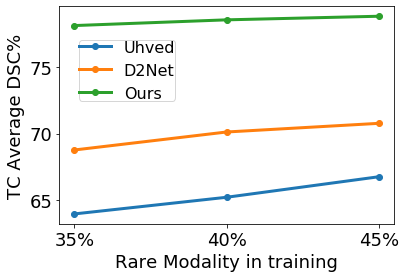}
    \label{figrare1supp}
  \end{subfigure}
  \begin{subfigure}{0.49\linewidth}
  \includegraphics[width=1.0\linewidth]{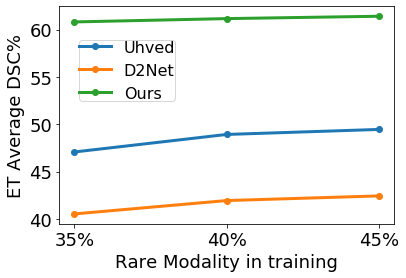}
    \label{figrare2supp}
  \end{subfigure}
  \caption{Ablation studies for varying \% of FLAIR in training.}
  \vspace{-.1in}
  \label{figraresupp}
\end{figure}

\begin{figure*}[hbtp]
\centering
    \begin{subfigure}{0.30\linewidth}
  \includegraphics[width=1\linewidth]{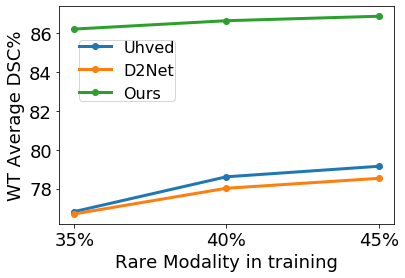}
    \label{figraret1c1}
  \end{subfigure}
  \begin{subfigure}{0.30\linewidth}
  \includegraphics[width=1\linewidth]{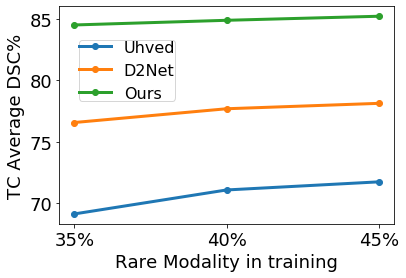}
    \label{figraret1c2}
  \end{subfigure}
  \begin{subfigure}{0.30\linewidth}
  \includegraphics[width=1\linewidth]{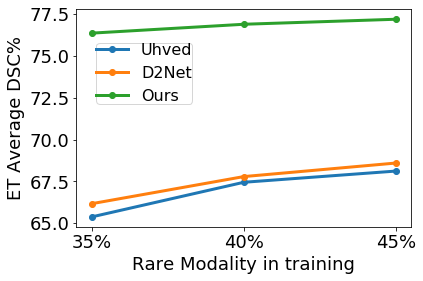}
    \label{figraret1c3}
  \end{subfigure}
  \caption{Ablation studies for varying \% of T1c in training.}
  \vspace{-.1in}
  \label{figraret1c}
\end{figure*}

\begin{figure}[hbtp]
  \vspace{-.05in}
  \includegraphics[width=1.0\linewidth]{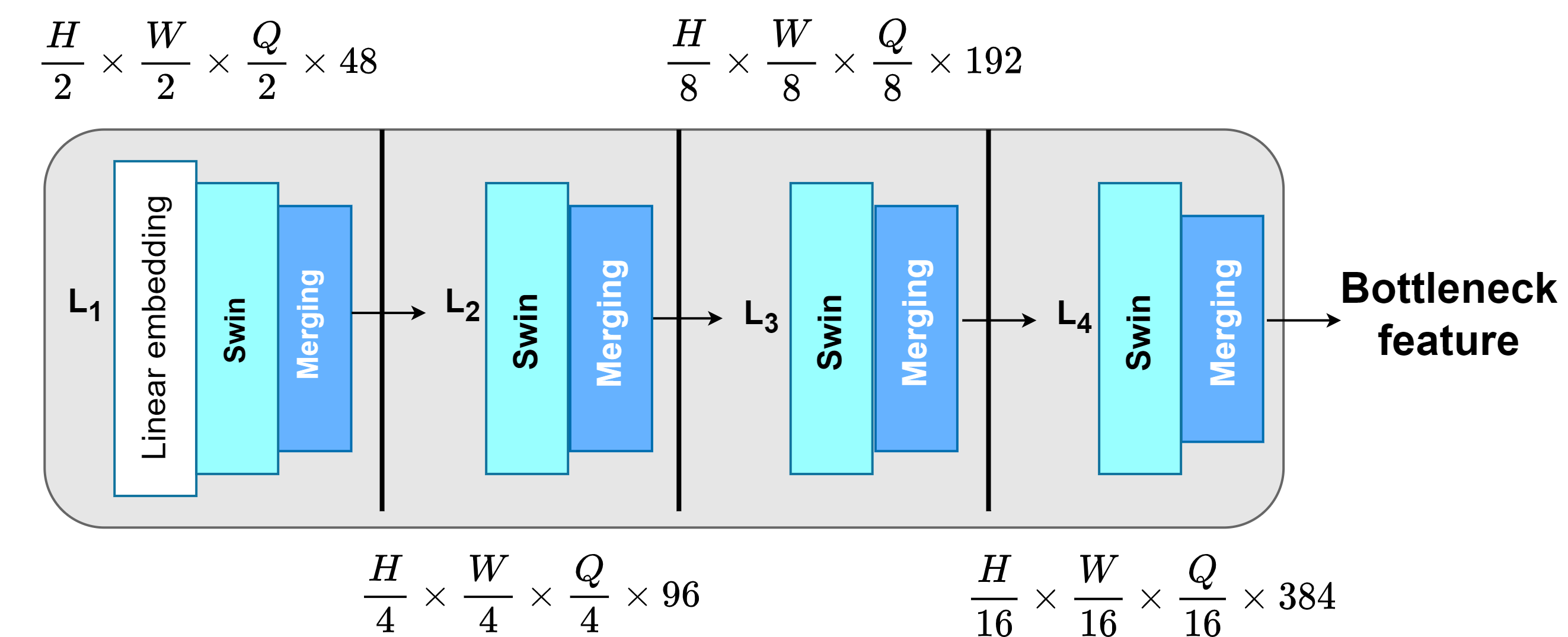}
    \vspace{-.1in}
     \caption{A schematic of the adopted Swin-UNETR encoder.}
      \vspace{.1in}
    \label{swin}
\end{figure}

In the main paper (Sec.~4.2, Fig.~7b) we have provided ablation results on the WT region by varying FLAIR proportion in training from 35\% to 45\%. Here we provide extensive results for the remaining two tumor regions (TC and ET).
Fig.~\ref{figraresupp} suggests that upon increasing FLAIR from 35\% to 45\%, our model's DSC gain (for both TC and ET) is much less when compared to that of U-HVED or D2Net. This demonstrates that our approach is not sensitive to presence of any particular modality. Similar conclusions can also be drawn when experiments are carried out keeping T1c as the rarely occurring modality instead of FLAIR. The results are presented in Tab.~\ref{tableabraresupp} and Fig.~\ref{figraret1c}.

\setlength{\tabcolsep}{2pt}
\begin{table}[t]
\fontsize{7}{9}\selectfont
		\centering
\begin{tabular}{c|c|c|c|c|c|c|c|c|c|c}
\hline
\multicolumn{1}{c|}{\multirow{4}{*}{M}} & FLAIR &  $\circ$ & $\circ$  & $\circ$  &  $\bullet$ &$\bullet$  & $\bullet$ & $\circ$ & $\bullet$ & \multirow{4}{*}{Avg} 
\\ 
\multicolumn{1}{c|}{}                   & T1    &  $\circ$  & $\circ$ & $\bullet$ &  $\circ$ & $\bullet$ &  $\circ$  & $\bullet$ & $\bullet$ &                            \\ 
\multicolumn{1}{c|}{}                   & T1c  &$\bullet$ &$\bullet$  &$\bullet$  & $\bullet$ & $\bullet$ & $\bullet$ & $\bullet$ & $\bullet$ &                             \\
\multicolumn{1}{c|}{}                   & T2    &  $\circ$ & $\bullet$ & $\circ$  & $\circ$ & $\circ$ & $\bullet$ & $\bullet$ & $\bullet$ &                            \\ \hline
\multirow{4}{*}{{WT}}                 &                     U-HVED    & 54.38 & 77.63 & 63.70 & 82.28 & 84.06 & 84.96 & 81.19 & 86.37 & 76.82                             \\ 
                                         & D2-Net     & 39.14 & 81.57 & 62.77 & 86.32 & 85.98 & 87.85 & 82.40 & 87.53 & 76.70                                                              \\ 
                                         & Ours     & \textbf{78.46}  & \textbf{85.71} & \textbf{78.83} & \textbf{89.22} & \textbf{89.64} & \textbf{90.08} & \textbf{87.19} & \textbf{90.57} & \textbf{86.21}                                   \\ 
\hline
\multirow{4}{*}{{TC}}                 &                     U-HVED     & 61.16 & 70.87 & 62.21 & 70.65 & 72.24 & 70.59 & 74.52 & 70.87 & 69.13                             \\ 
                                         & D2-Net    & 61.73 & 78.46 & 75.31 & 80.69 & 78.17 & 79.85 & 78.77 & 79.58 & 76.57                                                               \\ 
                                         & Ours     & \textbf{83.62}  & \textbf{84.79} & \textbf{83.56} & \textbf{84.18} & \textbf{84.35} & \textbf{83.88} & \textbf{85.80} & \textbf{85.94} & \textbf{84.51}                                   \\ 
\hline
\multirow{4}{*}{{ET}}                 &                     U-HVED     & 57.23 & 65.39 & 62.08 & 65.76 & 68.15 & 67.81 & 67.26 & 69.42 & 65.38                             \\ 
                                         & D2-Net     & 65.13 & 66.37 & 67.84 & 65.41 & 65.06 & 65.33 & 67.98 & 66.27 & 66.17                                                                 \\ 
                                         & Ours     & \textbf{77.02}  & \textbf{75.98} & \textbf{76.43} & \textbf{76.25} & \textbf{75.11} & \textbf{76.67} & \textbf{75.89} & \textbf{77.54} & \textbf{76.36}                                  \\ 
\hline
\end{tabular}
\caption{Ablation results for rare occurrence (35\%) of T1c in training. $\bullet$ for T1c in all combinations denote that T1c is always present in inference despite being rare on training.}
\vspace{.1in}
\label{tableabraresupp}

\end{table}

\section{Details on Feature Aggregation Module}
\label{fusioneq}
For a particular level $l$, a modality feature $\mathbf{F}^l_j \in \mathbb{R}^{C \times H \times W \times Q}$ includes $C$ channels and feature maps of size $H \times W \times Q$ where $j\in\{1,2,...,n\}$. The channels in these generated features are considered to encode relevant tumor-class specific information. Our fusion block exploits the correlation among available modality representations to develop a unified feature that best describes the tumor characteristics of a particular patient. First, the channel information $\gamma_j^l$ of a modality at level $l$ is preserved through the following equation:
\begin{equation}
\gamma_j^l = GAP(\mathbf{F}^l_j) = \frac{1}{H \times W \times Q} \sum_{h = 1}^{H} \sum_{w = 1}^{W} \sum_{q = 1}^{Q} \mathbf{F^l_j}(h,w,q),
\end{equation}
where $j\in\{1,2,...,n\}$ and $GAP$ denotes Global Average Pooling operation. Following this, we not only concatenate $\gamma_1^l,\gamma_2^l,...,\gamma_n^l$, but also impute zeros in the channel information of ($M-n$) missing modalities to form a resultant $M$-dimensional vector $\gamma^l$.
\begin{equation}
    \gamma^l = \gamma_1^l\oplus\gamma_2^l\oplus\gamma_3^l...\oplus\gamma_M^l,
\end{equation}
$\gamma^l$ is mapped to the channel weights
of $M$ modality features through a multi-layer perceptron $(MLP)$ and sigmoid activation function, $\sigma$.
\begin{equation}
    \Gamma^{l} =\sigma(MLP(\gamma^l)).
\end{equation}
Though $\Gamma^l$ contains $M$ scalar values, only the weights of $n$ available modalities are multiplied with their corresponding features. These weighted features are finally summed to obtain the fused representation $\mathbf{F}^{l}_{fused}$.
\begin{equation}
\vspace{-.05in}
   \mathbf{F}^{l}_{fused} = \sum_{j=1}^n\Gamma^l_j\mathbf{F}^{l}_j.
   \vspace{-.05in}
\end{equation}

\section{Fusion baselines and ablation}
\label{fusesupp}
\begin{figure*}[hbtp]
  \includegraphics[width=0.9\linewidth]{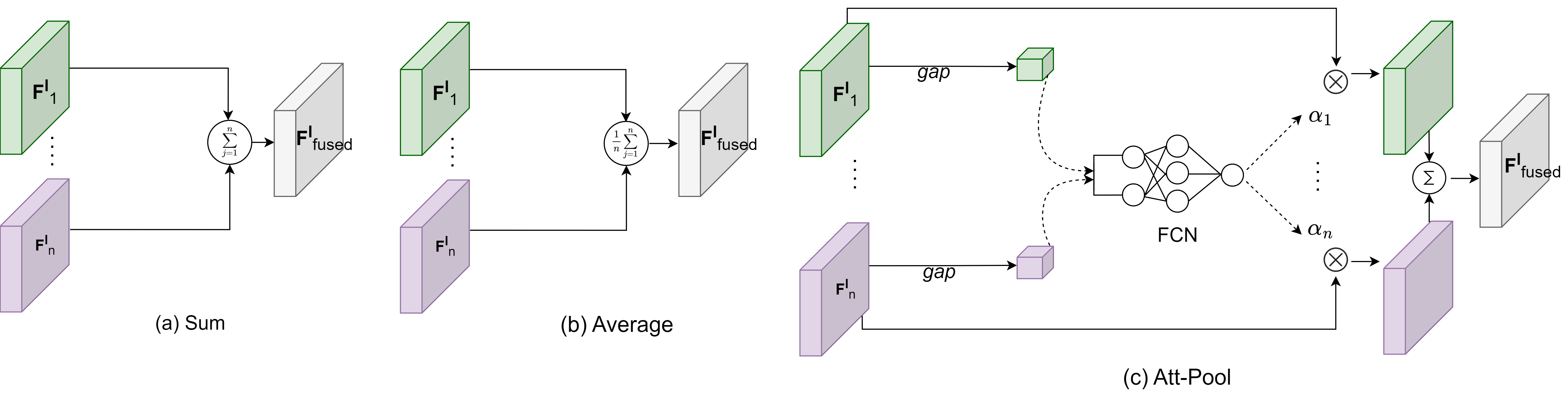}
  \vspace{-.15in}
    \caption{Fusion baselines}
  \vspace{.1in}
    \label{fusebaseline}
\end{figure*}

We design three baseline aggregation
modules to highlight the contribution of our fusion strategy. The architectures of the three fusion baselines, (a) Sum, (b) Average, and (c) Att-Pool are illustrated in Fig.~\ref{fusebaseline}. For the first two approaches, feature maps from the available modalities are summed or averaged along the channel dimension $C$ to obtain the fused feature. In the third approach, available modality features are individually passed through a Global Average Pooling (GAP) layer. The GAP outputs are fed to a Fully Connected Network (FCN) followed by a softmax activation function, producing the attention weights of each modality. Finally, attention-weighted summation of the original modality features gives rise to the fused feature. The ablation results are shown in Tab.~\ref{tablefuse}. Our feature aggregation block provides a better technique for dynamically learning from the heterogeneous input modalities, followed by inducing channel interaction among them. However, this plug-and-play fusion module is not a primary contribution and can be replaced by SOTA fusion techniques ~\cite{rfnet, robust}.



\setlength{\tabcolsep}{2.5pt}
\begin{table}[hbtp]
    \centering
    \scriptsize
\begin{tabular}{c|ccc}
\hline
\multicolumn{1}{c|}{\multirow{2}{*}{Methods}} & \multicolumn{3}{c}{Average DSC (\%)}                                       \\ \cline{2-4} 
\multicolumn{1}{c|}{}                         & \multicolumn{1}{c}{WT} & \multicolumn{1}{c}{TC} & ET \\ \hline
Sum                                       & \multicolumn{1}{c}{85.99}         &  \multicolumn{1}{l}{78.21}     &   60.85     \\
                                           Average     &  \multicolumn{1}{c}{86.14}         & \multicolumn{1}{c}{78.36}     & 61.30     \\
                                            Att-pool    &  \multicolumn{1}{c}{86.93}         & \multicolumn{1}{c}{79.07}     &  62.28      \\   
                                Ours       &  \multicolumn{1}{c}{\textbf{87.12}}         & \multicolumn{1}{c}{\textbf{79.12}}     & \textbf{62.53}        \\\hline
\end{tabular}
\caption{Ablation study on fusion.}
 \label{tablefuse}
\end{table}

\section{Qualitative comparison}
\label{segsupp}
\begin{figure}[hbtp]
  \includegraphics[width=1.0\linewidth]{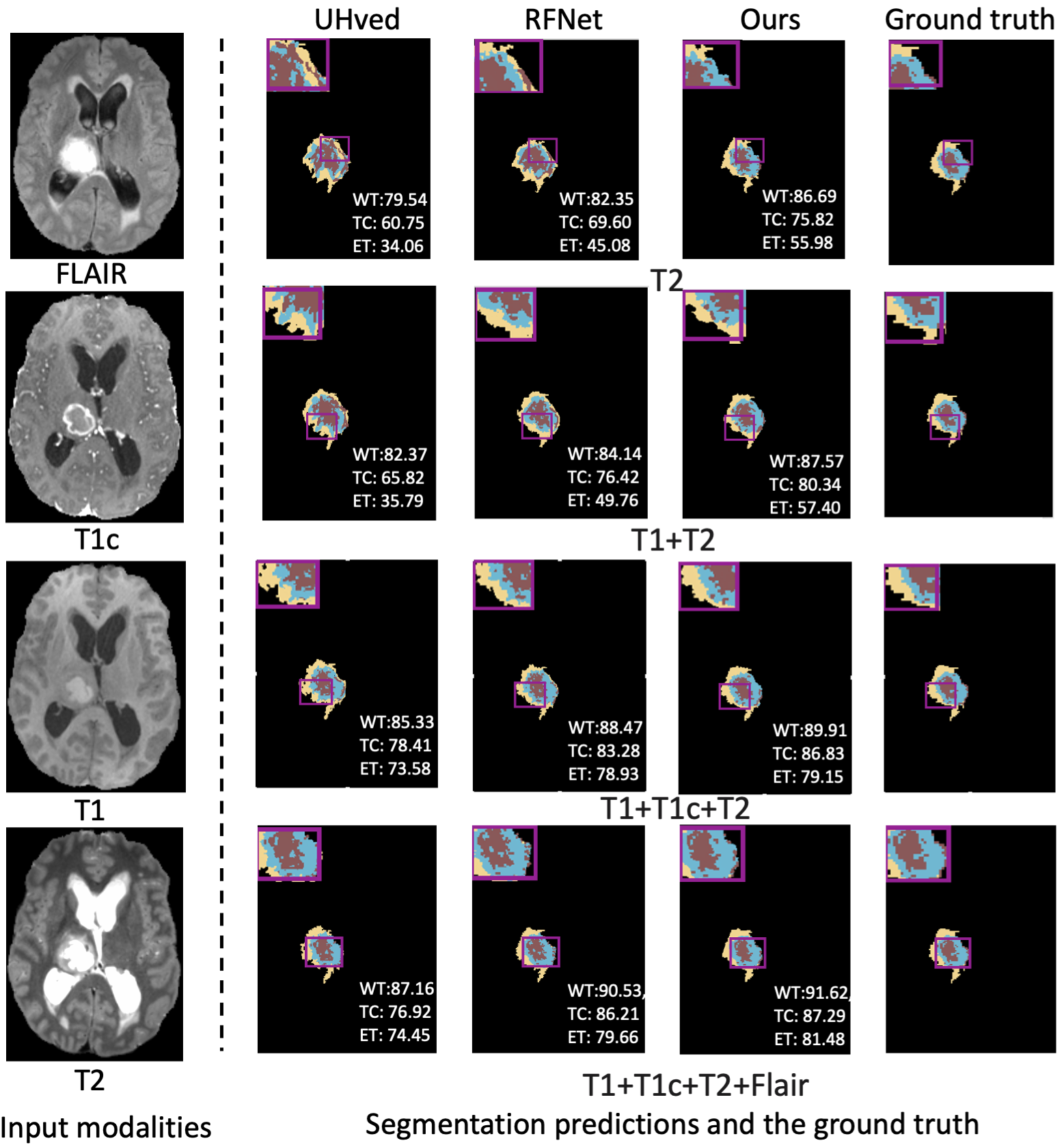}
  \vspace{-.15in}
    \caption{Qualitative comparisons with SOTA. Column 1: four MRI modalities. Column 2-4: segmentation maps predicted by three methods for different combinations of modalities. Column 5: Ground truth.}
  \vspace{.1in}
    \label{resultsupp}
\end{figure}

In Fig.~\ref{resultsupp} we visualize the segmentation masks predicted by U-HVED, RFNet, and our method from four combinations of modalities in the inference phase. Unlike other methods, our segmentations do not degrade sharply as additional modalities are dropped during inference. Even with single T2 or T1+T2 modalities, our model achieves higher DSC scores.

\section{Additional pre-processing and implementation details}
\label{implement}
As part of pre-processing, the organizers skull-stripped the volumes and interpolated them to an isotropic 1mm$^3$ resolution. For a given patient, the four sequences have been co-registered to the same anatomical template. Augmentations including random rotations, intensity shifts, and mirror flipping, are applied to the resized images. The foreground voxels within the brain are intensity-normalized to zero mean and unit standard deviation. We train our network using AdamW optimizer with an outer loop learning rate $\beta=5e-4$ for a maximum of 500 epochs. The two hyperparameters $\lambda_1$ and $\lambda_2$ in generator loss $\mathcal{L}_E$ are taken as 0.8 and 0.2, respectively. During training, $\mathcal{L}_{dis}$ is multiplied by 0.5
to prevent it from overpowering the generator.

Learning rate (LR: 5e-5), $\lambda_1$: 0.8, $\lambda_2$: 0.2, and scale of discriminator (Sc: 0.5) were selected based on the model performance. Results with different sets of parameters are shown in Tab.~\ref{tabaddimp}.

\setlength{\tabcolsep}{2pt}
\begin{table}[hbtp]
\centering
\scriptsize
\begin{tabular}{|c|c|}
\hline
LR   & WT DSC(\%) \\ \hline
5e-3 & 86.79      \\
5e-4 & 86.95      \\
5e-5 & 87.12     \\
\hline
\end{tabular}
\begin{tabular}{|c|c|}
\hline
$\lambda_1$, $\lambda_2$ & Avg DSC \% (WT, TC, ET) \\ \hline
0.9, 0.1 & 86.89, 78.94, 62.37     \\
0.8, 0.2 & 87.12, 79.12, 62.53     \\
0.7, 0.3 & 86.97, 78.83, 62.19   \\
\hline
\end{tabular}
\begin{tabular}{|c|c|}
\hline
Sc   & WT DSC(\%) \\ \hline
0.25 & 86.44      \\
0.5 & 87.12      \\
0.6 & 87.03     \\
\hline
\end{tabular}
\caption{Selection of experimental parameters}
\label{tabaddimp}
\end{table}

\section{Results on additional datasets}
\label{nonbrats}
We show the segmentation results (Tab.~\ref{xyz}, \ref{abc}) on two additional datasets not in the BRATS cohort. The first dataset, $D_1$ contains 4 MRI modalities for 80 patients. For this dataset, we segment brain glioma tumors into 3 regions (WT, TC, ET). Another dataset, $D_2$ contains 1 MRI modality (FLAIR) and 1 CT modality for 85 patients with metastatic brain tumors as the segmentation targets. Unlike the solitary brain tumors studied in the other datasets, multiple distinct metastatic targets can occur at multiple locations within a patient's brain for $D_2$. 

\setlength{\tabcolsep}{1pt}
\begin{table}[ht]
  \begin{minipage}{0.22\textwidth}
  \footnotesize
    \centering
    \scriptsize
\begin{tabular}{c|ccc}
\hline
\multicolumn{1}{c|}{\multirow{2}{*}{Methods}} & \multicolumn{3}{c}{Average DSC (\%)}                                       \\ \cline{2-4} 
\multicolumn{1}{c|}{}                         & \multicolumn{1}{c}{WT} & \multicolumn{1}{c}{TC} & ET \\ \hline
U-HVED [14]                                       & \multicolumn{1}{c}{75.37}         &  \multicolumn{1}{l}{60.29}     &   47.52     \\
                                           RFnet [13]     &  \multicolumn{1}{c}{81.04}         & \multicolumn{1}{c}{72.15}     & 53.22     \\
                                            mmFormer [55]   &  \multicolumn{1}{c}{81.73}         & \multicolumn{1}{c}{71.31}     &  51.49      \\   
                                Ours       &  \multicolumn{1}{c}{\textbf{82.53}}         & \multicolumn{1}{c}{\textbf{74.26}}     & \textbf{56.13}        \\\hline
\end{tabular}
\caption{Results on $D_1$.}
 \label{xyz}
  \end{minipage}
  \hfill
  \begin{minipage}{0.22\textwidth}
  \footnotesize
    \centering
    \scriptsize
\begin{tabular}{c|c|c|c|c}
\hline
FLAIR & $\bullet$ & $\circ$ & $\bullet$ & \multirow{2}{*}{Avg DSC} 
\\
CT    & $\circ$ & $\bullet$ & $\bullet$ &                            \\ \hline
U-HVED    &  49.93 & 46.20 & 48.59 & 48.24         \\ 
                                         RFNet     & 53.62 & 51.37 & 53.16 & 52.71
                                         \\ 
                                         mmFormer     & 54.88 & 52.85 & 54.63 & 54.12 \\ 
                                         Ours     &  \textbf{55.19} &  \textbf{53.27} & \textbf{55.06} & \textbf{54.50}                                   \\ 
\hline
\end{tabular}
\caption{Results on $D_2$.}
 \label{abc}
  \end{minipage}
  \hfill
  \vspace{-.15in}
\end{table}

\end{document}